\documentclass{article} 
\usepackage{iclr2026_conference,times}
\usepackage{xcolor}

\usepackage{amsmath,amsfonts,bm}

\def\eqref#1{equation~\ref{#1}}

\def\1{\bm{1}}

\DeclareMathAlphabet{\mathsfit}{\encodingdefault}{\sfdefault}{m}{sl}
\SetMathAlphabet{\mathsfit}{bold}{\encodingdefault}{\sfdefault}{bx}{n}

\usepackage{wrapfig}
\usepackage{hyperref}
\usepackage{url}
\usepackage{bbm}
\usepackage[skins]{tcolorbox}
\usepackage[capitalize, noabbrev]{cleveref}
\usepackage[font=small,labelfont=bf]{caption}
\usepackage{lipsum}
\usepackage{listings}
\usepackage{placeins}
\usepackage{booktabs}
\usepackage{graphicx}   
\usepackage{array}      
\usepackage{multirow}   
\usepackage{array} 
\usepackage{tabularx}
\usepackage{wrapfig}
\usepackage{subcaption}
\definecolor{darkgreen}{RGB}{0,100,0} 

\definecolor{gtgreen}{HTML}{006400}
\definecolor{deepred}{rgb}{0.7,0.0,0.0}
\definecolor{mildyellow}{rgb}{0.92,0.9,0.88}
\definecolor{steelblue}{rgb}{0.27,0.51,0.71}
\definecolor{darkcyan}{rgb}{0.0,0.55,0.55}

\newcommand{\bfx}{\mathbf{x}}
\newcommand{\ols}[1]{\mskip.5\thinmuskip\overline{\mskip-.5\thinmuskip {#1} \mskip-.5\thinmuskip}\mskip.5\thinmuskip}

\usepackage{booktabs,multirow}
\usepackage{xcolor}
\definecolor{codegreen}{rgb}{0,0.6,0}
\definecolor{codegray}{rgb}{0.5,0.5,0.5}
\definecolor{codepurple}{rgb}{0.58,0,0.82}
\definecolor{backcolour}{rgb}{0.98,0.98,0.96}

\lstdefinestyle{pythonstyle}{
    commentstyle=\color{codegreen},
    keywordstyle=\color{magenta},
    numberstyle=\tiny\color{codegray},
    stringstyle=\color{codepurple},
    basicstyle=\ttfamily\small,
    breakatwhitespace=true,
    breaklines=true,
    captionpos=b,
    keepspaces=true,
    numbers=left,
    numbersep=5pt,
    showspaces=false,
    showstringspaces=false,
    showtabs=false,
    tabsize=4,
    language=Python
}
\usepackage{tikz}
\pgfdeclarehorizontalshading{rainbow}{100bp}{
  color(10bp)=(red);
  color(40bp)=(yellow);
  color(50bp)=(green);
  color(60bp)=(blue);
  color(80bp)=(violet)
}

\title{Rainbow Padding: Mitigating Early Termination in Instruction-Tuned Diffusion LLMs}

\author{
\!Bumjun Kim$^{1}$\thanks{Equal Contribution.} \hspace{0.5em} Dongjae Jeon$^{2*}$ \hspace{0.5em} Dueun Kim$^{1*}$ \hspace{0.5em} \\ 
\textbf{Wonje Jeung}$^{1}$  \hspace{0.6em} \textbf{Albert No}$^{1}$\thanks{Corresponding Author.} \vspace{0.3em}\\ { $^1$Artificial Intelligence Department of Yonsei University}\\
{ $^2$Computer Science Department of Yonsei University}
}

\newcommand{\mask}{\texttt{[M]}}
\newcommand{\llada}{\text{LLaDA}}
\newcommand{\dream}{\text{Dream}}
\newcommand{\mxlen}{\texttt{max\_length}}

\iclrfinalcopy 
\begin{document}

\maketitle

\begin{abstract}
Diffusion large language models (dLLMs) have emerged as a promising alternative to autoregressive models, offering flexible generation orders and strong performance on complex reasoning tasks. 
However, instruction-tuned dLLMs exhibit a critical vulnerability we term \texttt{<eos>} overflow: as allocated sequence length increases, responses paradoxically become shorter, collapsing into early termination or degenerating into streams of \texttt{<eos>} tokens. 
Although noticed in practice, this issue has not been systematically analyzed. 
We trace its root cause to the dual role of \texttt{<eos>} as both termination and padding, which concentrates probability mass on \texttt{<eos>} at later positions and propagates backward to trigger early termination. 
To address this, we introduce Rainbow Padding, a simple remedy that replaces repeated \texttt{<eos>} placeholders with a repeating cycle of distinct padding tokens, distributing probability mass and breaking \texttt{<eos>} dominance. 
Experiments show that Rainbow Padding substantially improves length robustness and output quality, with as few as seven padding tokens sufficient to prevent early termination. 
Moreover, the method integrates efficiently into existing instruction-tuned models: LoRA fine-tuning for a single epoch on minimal data yields significant improvements, making this solution highly practical.
The code is publicly available at ~\url{https://github.com/quasar529/rainbow-padding}.
\end{abstract}

\section{Introduction}

\begin{wrapfigure}{r}{0.54\textwidth}
\vskip -.2in
    \centering
    \includegraphics[width=\linewidth]{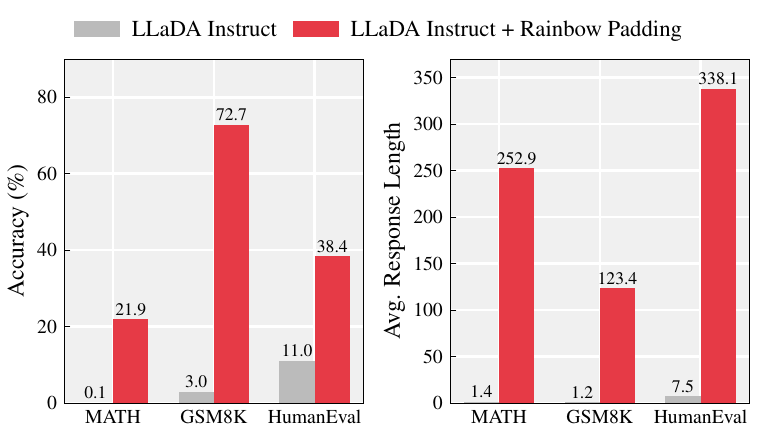}
    \vskip -.06in
    \caption{
Performance comparison of LLaDA-Instruct with and without Rainbow Padding.
Standard LLaDA\protect\footnotemark produces overly short responses at moderate generation budgets (\texttt{max\_length} = 1024), resulting in significant accuracy degradation.
Adapting with Rainbow Padding yields substantial performance gains.
    }
\label{fig:intro}
\vskip -.3in
\end{wrapfigure}

\footnotetext{Throughout this paper, LLaDA and Dream denote the instruction-tuned models \texttt{LLaDA-8B-Instruct} and \texttt{Dream-v0-Instruct-7B}, unless stated otherwise.}

Discrete Diffusion large language models (dLLMs)~\citep{nie2025llada, zhu2025llada1_5, ye2025dream, labs2025mercury, gemini2025diffusion} have recently emerged as a promising alternative to traditional autoregressive LLMs. Unlike autoregressive models, which generate strictly left-to-right, dLLMs allow tokens to be generated in \emph{any order}, while maintaining global consistency through a diffusion-style denoising process. This flexible decoding has been linked to stronger multi-step reasoning and planning ability~\citep{ye2024sudoku, kim2025train}, with benefits that persist at scale~\citep{nie2024scaling}. As a result, dLLMs are increasingly positioned as a viable paradigm for foundation models.

However, current instruction-tuned dLLMs suffer from a critical reliability issue. When users allocate longer generation budgets (\texttt{max\_length}), these models often produce \emph{shorter} responses: terminating early or degenerating into streams of \texttt{<eos>} tokens. We refer to this failure mode as \texttt{<eos>} overflow. This paradox—where giving the model more space yields worse results—has been observed in practice~\citep{nie2025llada, zhu2025llada1_5} but has not been systematically analyzed. It's impact is substantial: performance on reasoning and coding tasks deteriorates even at moderate sequence lengths, undermining the utility of dLLMs in real-world instruction-following scenarios. 

We trace the root cause of this issue to a design flaw in the instruction-tuning process.
Current pipelines pad variable-length sequences with the \texttt{<eos>} token, thereby assigning it a dual role—as both the legitimate end-of-sequence marker and a placeholder for unused positions.
This conflation introduces a strong positional bias: \texttt{<eos>} appears disproportionately at later positions. 
When used with widely adopted probability-based decoding strategies~\citep{chang2022maskgit, kim2025train, ye2025dream, ben2025accelerated}, which typically prioritize the most confidently predicted tokens, \texttt{<eos>} is often selected prematurely.
Once sampled at the tail, \texttt{<eos>} predictions propagate backward through the sequence, resulting in the overflow effect and early termination.

To address this failure, we introduce \emph{Rainbow Padding}, a simple yet effective modification to the padding scheme. 
Rather than repeating \texttt{<eos>} throughout the tail, we reserve a single \texttt{<eos>} to mark the true end of the sequence and fill the remainder with a cyclic palette of distinct padding tokens. 
This design has two key effects: it decouples termination from padding, ensuring that \texttt{<eos>} is learned only as a proper stopping symbol; and it distributes probability mass across multiple tokens, preventing any single padding token from dominating. 
The deterministic cycle is easy to learn and provides a weak structural signal of length without hindering the model’s ability to learn meaningful contextual dependencies during training.

Rainbow Padding can be adopted with only a brief fine-tuning phase, even for already instruction-tuned models. 
Despite its simplicity, it effectively eliminates \texttt{<eos>} overflow, restoring length robustness and significantly improving performance on mathematical reasoning, code generation, and general instruction-following tasks. 
Figure~\ref{fig:intro} illustrates the effect: as \texttt{max\_length} increases, baseline dLLM (\llada) collapses into short answers, whereas Rainbow Padding restores appropriate response length and accuracy (see Sections~\ref{sec:experiment}--\ref{sec:extra_analysis}). Unlike heuristic fixes—such as manually suppressing \texttt{<eos>} confidence or enforcing semi-autoregressive block decoding with sensitive hyperparameters—Rainbow Padding resolves the issue as an inherent property of the model, achieved through a simple change in the padding scheme. 
It is architecture-agnostic, dataset-agnostic, robust to decoding strategies, and lightweight to deploy, making it a practical standard for robust instruction-tuning of dLLMs.

Our contributions are summarized as follows:
\begin{itemize}
    \item We define and measure \texttt{<eos>} overflow—a failure mode unique to dLLMs—at both the task and token levels, demonstrating its severe impact on instruction-following and reasoning benchmarks.
    \item We analyze how confidence-based decoding amplifies padding-induced bias and show how a structured, cyclic padding scheme breaks the overflow cascade.
    \item We propose Rainbow Padding, a cyclic multi-pad scheme that restores stable length control with minimal training overhead. We validate its effectiveness through controlled ablations, showing that as few as seven distinct padding tokens are sufficient to resolve the issue, with robustness across a variety of decoding strategies.

\end{itemize}

\section{Preliminaries}
\label{sec:prelim}
\paragraph{Diffusion Language Modeling.}
Diffusion models approximate complex data distributions via a latent variable framework that consists of a forward noising process and a reverse denoising process~\citep{sohl2015diffusion, song2021scorebased, ho2020ddpm}. 
While initially proposed for continuous domains such as images, recent work has extended diffusion to the discrete setting, showing strong promise for language modeling~\citep{austin2021d3pm, nie2024mdm}.

Among several formulations, the dominant approach for discrete text generation is \emph{masked diffusion}.
Let $\mathbf{x}=(x_{1},x_{2},\ldots,x_{L})$ be a sequence of length $L$ from vocabulary $\mathcal{V}$ augmented with a special mask token \mask. 
The forward process is an absorbing-state Markov chain where each token can only be corrupted into \mask. 
Formally, for each position $i$, let $M_i\in \{0,1\}$ be a Bernoulli indicator such that $M_i=1$ if $x_i$ is replaced by \mask. 
Denote $\mathbf{M}=\{i:M_i=1\}$ as the masked indices and $\ols{\mathbf{M}}=\{i:M_i=0\}$ as the unmasked indices. 
The learning task is then to model the clean conditional distribution
\[
p_\theta(x_i \,|\, \bfx_{\ols{\mathbf{M}}}), \quad i\in\mathbf{M},
\]
that is, to correctly \emph{guess the masked tokens given the unmasked partial sequence}. 

Training proceeds by first sampling a corruption rate $\lambda\sim U(0,1)$ and then masking each position independently, $M_i \sim \text{Bern}(\lambda)$. 
The model parameters are optimized by minimizing the cross-entropy objective
\begin{equation}\nonumber
\mathcal{L}(\theta) = -\mathbb{E}_{\bfx,\,\lambda,\,\ols{\mathbf M}}\left[\frac{1}{\lambda}\sum_{i=1}^L M_i \log p_\theta(x_i \,|\, \bfx_{\ols{\mathbf{M}}})\right],
\label{eq:masked-ce}
\end{equation}
where $1/\lambda$ normalizes for the expected fraction of masked tokens.

This masked formulation is both simple and effective: it avoids the instability of more complex transition designs, supports continuous-time parameterizations~\citep{lou2024sedd, sahoo2024mdlm, shi2024md4, gong2025scaling}, and allows weight sharing across timesteps, yielding a time-independent estimator~\citep{ou2024radd}. 
As a result, state-of-the-art open-source diffusion large language models (dLLMs) such as LLaDA~\citep{nie2025llada} and Dream~\citep{ye2025dream} adopt masked diffusion as their core framework, typically built on Transformer encoder architectures~\citep{vaswani2017transformer, peebles2023dit}. 
Throughout this paper, we will use this masked diffusion formulation to analyze and improve instruction-tuned dLLMs.

\paragraph{Decoding Strategies in Diffusion Language Models.}
A central advantage of dLLMs over autoregressive models is their flexible \emph{any-order decoding}: 
rather than being locked into a left-to-right order, the model adaptively chooses which masked positions to reveal first. 
This \emph{adaptive decoding} is what gives dLLMs their potential for tasks such as planning,
constraint satisfaction, or coarse-to-fine generation, 
but it also makes them highly sensitive to the choice of decoding policy. 
Because training is imperfect, different unmasking orders induce different distributions 
and can even change failure modes~\citep{chang2022maskgit, zheng2024gumbel, kim2025train, ye2025dream}.

Several simple heuristics are widely used:
\begin{itemize}
  \item \textbf{Confidence:} select the position with the highest peak probability.
  \item \textbf{Margin:} select the position with the largest gap between the top-1 and top-2 probabilities.
  \item \textbf{Entropy:} select the position with the lowest predictive entropy.
\end{itemize}

These adaptive strategies all aim to reveal ``easy'' tokens first and use them as anchors for harder positions. 
In this work we adopt the confidence-based strategy, as it is the most widely used and also the most sensitive to early termination.

\begin{wrapfigure}{r}{0.39\textwidth}
\vskip -.17in
    \centering
    \includegraphics[width=\linewidth]{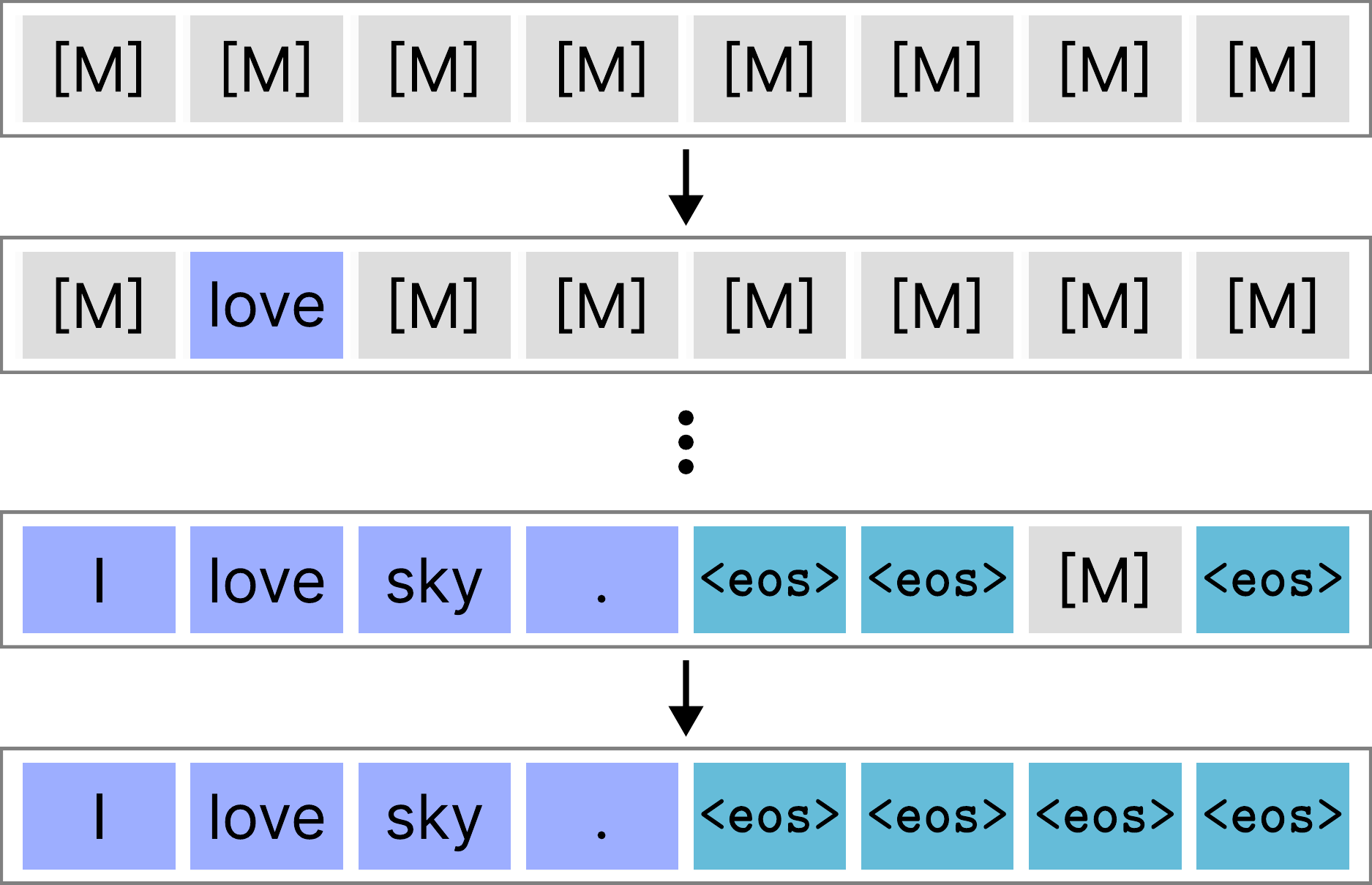}
    \vskip -.07in
    \caption{
Inference in dLLMs: to control response length, dLLMs must learn explicit padding tokens-unlike AR models. Current dLLMs use \texttt{<eos>} as padding.}
    \label{fig:illust_dd_pad}
\vskip -.3in
\end{wrapfigure}

\paragraph{Instruction-tuning in Diffusion Language Models.}
Instruction-tuning (IT) is essential to the success of large language models (LLMs), as it aligns pretrained models 
with user instructions and enables strong zero-shot generalization across downstream tasks~\citep{wei2021it1, zhang2023it2}. 
During IT, the model is fine-tuned on batches of \texttt{(instruction, response)} pairs of varying lengths. 
To enable efficient batching, shorter responses are padded to a fixed length by appending padding tokens. 
For autoregressive (AR) LLMs, this convention is harmless:
the model learns to stop at the \texttt{<eos>} token and excludes padding tokens from the training objective.

In contrast, diffusion LLMs (dLLMs) operate on the entire fixed-length sequence at every decoding step. 
Here, using padding tokens can be problematic: the model repeatedly observes padding tokens in trailing positions and incorporates them into both attention and the loss function. 
Current dLLMs use \texttt{<eos>} as padding for convenience (see~\cref{fig:illust_dd_pad}). 
As a result, the \texttt{<eos>} token is heavily overexposed, leading to biased termination probabilities that interact strongly with adaptive decoding strategies. 
This conflation of padding and termination introduces a critical vulnerability for dLLMs, one we examine in detail in the following section.

\section{Early Termination in Instruction-Tuned dLLMs}\label{sec:curse}

\begin{wrapfigure}[23]{r}{0.36\textwidth}
\vskip -.16in
    \centering

    \begin{minipage}{\linewidth}
        \includegraphics[width=\linewidth]{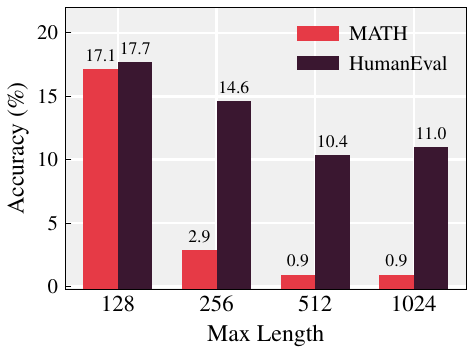}
        \vspace{-1.9em}
    \captionsetup{type=figure,width=.95\linewidth,singlelinecheck=false}
        \captionof{figure}{Accuracy of two tasks for LLaDA with varying \texttt{max\_length}.}
        \label{fig:acc_comp_len}
    \end{minipage}
\vskip +.23in
    \begin{minipage}{\linewidth}
        \includegraphics[width=\linewidth]{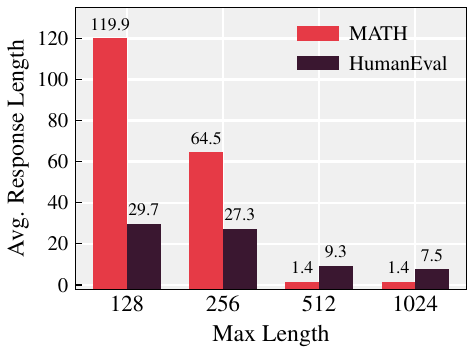}
        \vspace{-1.9em}
        \captionsetup{type=figure,width=.95\linewidth,singlelinecheck=false}
        \captionof{figure}{Average response length of LLaDA with varying \texttt{max\_length}.}
        \label{fig:len_comp_len}
        \vspace{1.9em}
    \end{minipage}
\end{wrapfigure}

\paragraph{Fixed generation length.}  
Unlike autoregressive (AR) models, diffusion large language models (dLLMs) require a fixed generation length (\texttt{max\_length}) that must be specified in advance. 
If the allocated length is too short, responses may be truncated; if it is longer than needed, the remaining positions are filled with padding tokens. 
In principle, well-trained dLLMs should remain stable as long as \texttt{max\_length} exceeds the minimum required response length.

\paragraph{Paradoxical degradation.}  
Surprisingly, we find that the performance of instruction-tuned dLLMs degrades sharply as \texttt{max\_length} increases. As shown in \cref{fig:acc_comp_len}, performance of \llada\,\,drops substantially at 512 and collapses further at 1024 tokens. 
This behavior is counterintuitive: many benchmark tasks require long and detailed responses, so allocating more tokens should, if anything, improve quality. 
Moreover, lengths of 512 or 1024 tokens are modest by modern LLM standards, making this degradation particularly striking.

\paragraph{Shorter responses with longer allocation.}  
Closer inspection reveals that longer \texttt{max\_length} allocations actually lead to \emph{shorter} responses. 
Figure~\ref{fig:len_comp_len} shows that average response length, measured up to the first \texttt{<eos>} token, decreases as \texttt{max\_length} increases. 
In extreme cases, models produce almost no content, collapsing into degenerate streams of \texttt{<eos>} tokens (Figure~\ref{fig:eos_overflow}). 
We refer to this phenomenon as \texttt{<eos>} overflow: paradoxically, allocating more space makes the model terminate earlier.

\begin{figure}[!h]
\centering
    \fcolorbox{deepred}{mildyellow}{
    \begin{minipage}{0.98\columnwidth}
    \textcolor{gray}{[Question]} Can you make a python function for factorial using recursion? \par
    \vspace{4pt}\hrule\vspace{4pt}
    \textcolor{gray}{[Answer: \texttt{max\_length} 128]} Sure, 
    
    def factorial(n):
    
    \quad if n==1:
    
    \qquad return 1
    
    \quad return n*factorial(n-1) \texttt{<eos>}\texttt{<eos>}\ldots
    
    \vspace{4pt}\hrule\vspace{4pt}
    \textcolor{gray}{[Answer: \texttt{max\_length} 1024]} Sure, \texttt{<eos>}\texttt{<eos>}\texttt{<eos>}\texttt{<eos>}\texttt{<eos>}\texttt{<eos>}\texttt{<eos>}\ldots
    \end{minipage}
    }
    \caption{
Illustrative example of \texttt{<eos>} overflow in LLaDA. With small \mxlen, the model produces valid answers, but as \mxlen~increases, it fills most positions with \texttt{<eos>}.}
    \label{fig:eos_overflow}
\end{figure}

\paragraph{Root cause: dual use of \texttt{<eos>}.}  
This failure arises from instruction-tuning practices. 
Current dLLMs reuse the same \texttt{<eos>} token both to mark the natural end of a response and to fill unused positions as padding. 
This dual use introduces two issues. 
First, the model cannot reliably distinguish whether a given \texttt{<eos>} indicates a true stopping point or padding, weakening its ability to learn correct termination. 
Second, because training batches contain responses of varying lengths, later positions are disproportionately filled with \texttt{<eos>}. 
Under masked cross-entropy training, the model’s predictions align with empirical token frequencies:
\[
\mathbb{E}_{\mathbf{x},\,\ols{\mathbf{M}}}\big[p_\theta(x_i=\texttt{<eos>}\,|\,\bfx_{\ols{\mathbf{M}}})\big]
\;\approx\; \Pr_{\mathbf{x}}[x_i=\texttt{<eos>}],
\]
and since $\Pr[x_i=\texttt{<eos>}] \to 1$ as $i$ approaches the maximum length $L$, the model learns excessively high priors for \texttt{<eos>} at the tail.

\begin{figure}[!htbp]
  \centering
  \begin{minipage}{0.95\textwidth}
    \centering
    \begin{subfigure}{0.48\linewidth}
      \includegraphics[width=\linewidth]{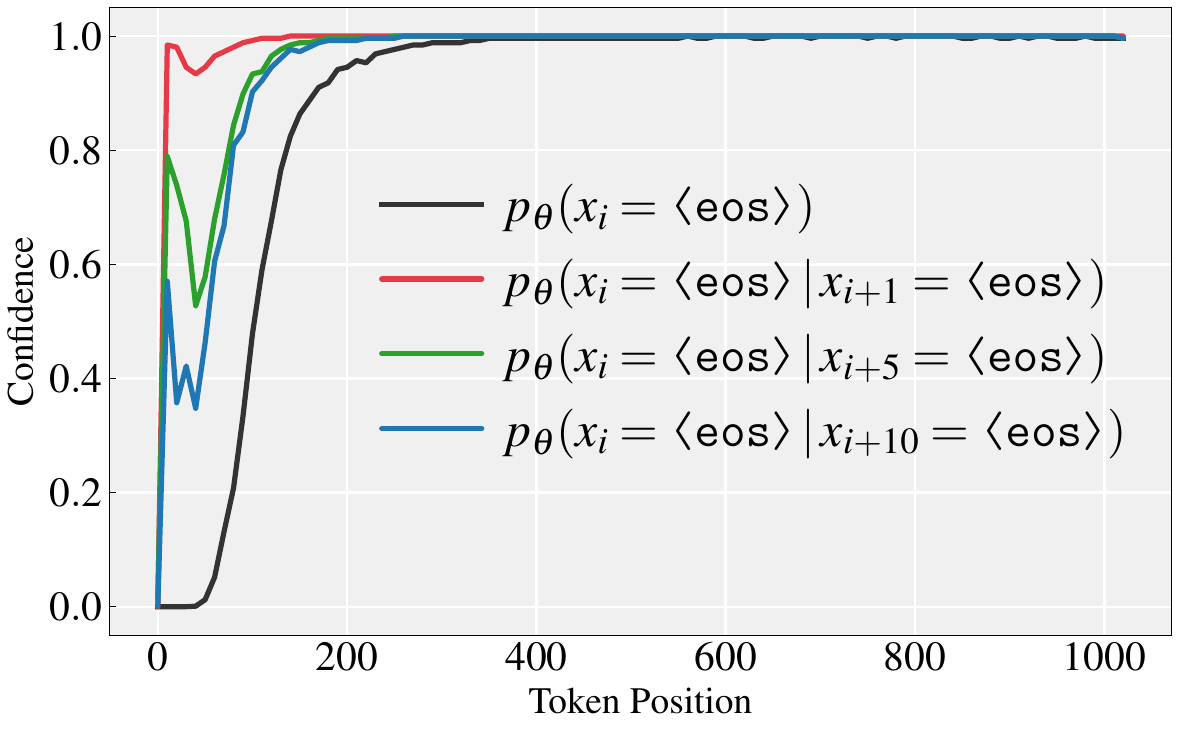}
      \caption{
\texttt{<eos>} confidence at each position with \texttt{max\_length} = 1024. 
\texttt{<eos>} at later positions create cascading bias affecting earlier positions.
      }
      \label{fig:eos_confidence}
    \end{subfigure}\hfill
    \begin{subfigure}{0.49\linewidth}
      \includegraphics[width=0.98\linewidth]{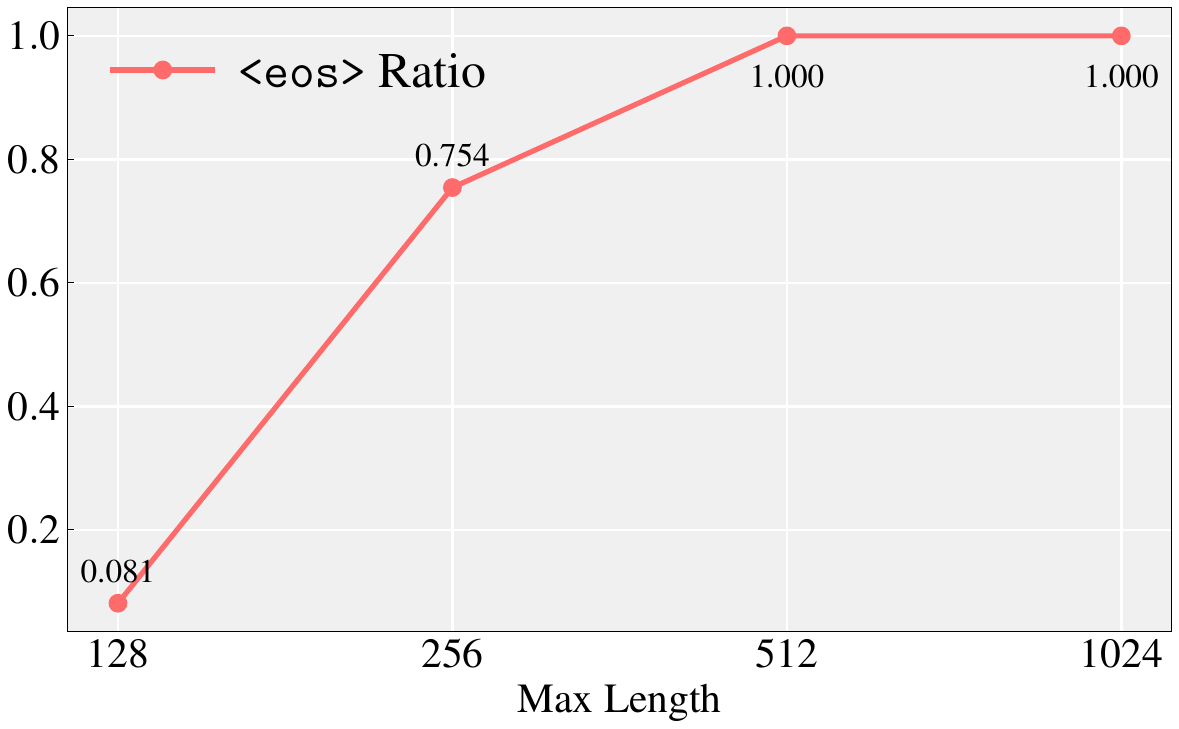}
      \caption{
Average \texttt{<eos>} token ratio among the first 50\% of unmasked tokens across different \mxlen~using confidence-based decoding.
      }
      \label{fig:unmasking_ratio}
    \end{subfigure}
    \caption{
Excessive \texttt{<eos>} generation in LLaDA on MATH.
    }
    \label{fig:llada_math}
  \end{minipage}%
  \vskip-.15in
\end{figure}

\paragraph{\texttt{<eos>} overflow.}  
This bias is further amplified by adaptive decoding. 
\Cref{fig:eos_confidence} shows that the probability of predicting \texttt{<eos>} rises steeply toward the sequence end, approaching 1.0 even before generation begins. 
Once a tail position is sampled as \texttt{<eos>} due to this high probability, earlier positions are also biased toward the same outcome, as quantified by
\[
p_\theta(x_i=\texttt{<eos>} \,|\, x_{i+k}=\texttt{<eos>}),
\]
which increases sharply even for positions 10 tokens earlier.
This cascading effect, \texttt{<eos>} \emph{overflow}, causes termination probabilities to propagate backward from the tail, ultimately collapsing the response. 
Figure~\ref{fig:unmasking_ratio} confirms this dynamic: longer \texttt{max\_length} allocations lead to substantially higher fractions of \texttt{<eos>} tokens unmasked within the first 50\% of decoding steps, preventing the generation of richer content.

\begin{wrapfigure}{r}{0.36\textwidth}
    \vspace{-1.3em}
    \centering
    \includegraphics[width=\linewidth]{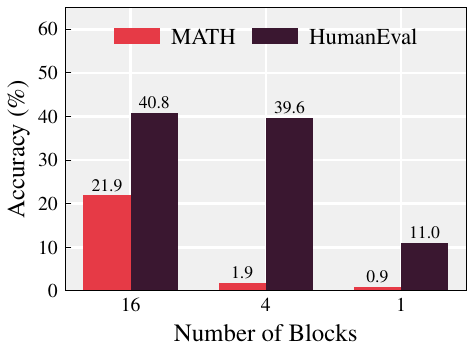}
    \vskip -0.03in
    \caption{
Performance of LLaDA with different block numbers during semi-autoregressive block decoding at \texttt{max\_length} = 1024.
}
\label{fig:acc_block_size}
\vskip -0.2in
\end{wrapfigure}

\paragraph{Heuristic fixes.}  
Although excessive \texttt{<eos>} generation at longer lengths has not been formally analyzed, several works have attempted to mitigate it with ad hoc strategies. 
For example,~\citet{zhu2025llada1_5} manually suppressed \texttt{<eos>} probabilities during decoding and reported modest gains. 
However, dampening confidence in this token often causes the model to overshoot the true response length, leading to failed termination and repetitive outputs (e.g., solving the same problem multiple times), which degrades quality (refer to Appendix~\ref{app:conf_lowering} for detail).

Another approach, adopted by \llada, is block-wise decoding in a semi-autoregressive manner. 
Here the sequence is partitioned into contiguous blocks, and later blocks remain masked until earlier ones are fully generated. 
This prevents premature \texttt{<eos>} predictions at the tail but at the cost of enforcing a sequential schedule, introducing a mismatch between training and inference. 
The restriction undermines a core strength of diffusion models—the ability to unmask tokens in arbitrary order with bidirectional context—which is critical for tasks requiring multi-step reasoning or subgoal planning, where dLLMs often outperform AR models~\citep{ye2024sudoku, kim2025train, ye2025dream}.

Block-wise decoding also introduces a sensitive hyperparameter: block number. 
In practice, different block numbers are chosen across benchmarks without a principled rule, and performance varies substantially with this choice (\cref{fig:acc_block_size}). 
This sensitivity highlights both the instability of heuristic fixes and the need for a fundamental solution to \texttt{<eos>} overflow.

\section{Rainbow Padding: A Simple Remedy for Early Termination}

\label{sec:method}

\begin{figure*}[!t]
    \centering
    \includegraphics[width=\textwidth, trim = 0cm 0cm 0cm 0cm, clip]{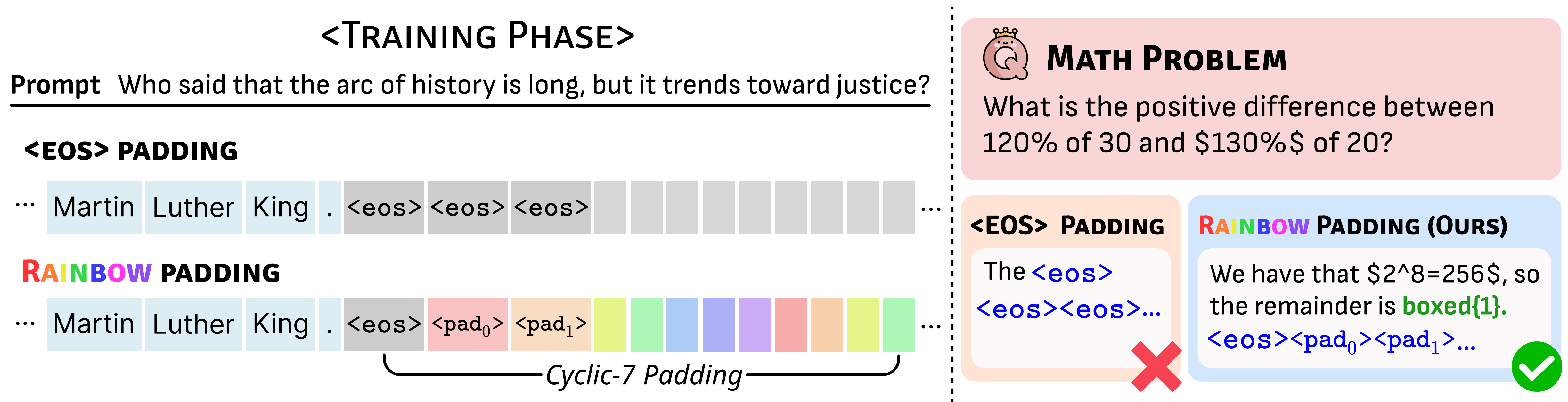}
    \caption{
\textbf{Overview of Rainbow Padding}.
(Left) During training, distinct \texttt{<pad>} tokens are arranged in a cyclic pattern, contrasting with current dLLMs that use \texttt{<eos>} padding. (Right) Models trained with \texttt{<eos>} padding suffer from \texttt{<eos>} overflow, which Rainbow Padding resolves.
    }
    \label{fig:rainbow_main}
\end{figure*}

In the previous section, we explain that excessive \texttt{<eos>} tokens at sequence ends during instruction-tuning cause \texttt{<eos>} overflow. 
Replacing \texttt{<eos>} in padding regions with a single \texttt{<pad>} token is insufficient, as it reintroduces the same concentration of probability mass that causes overflow.

\paragraph{Design.}  
We introduce \emph{Rainbow Padding} to resolve this issue.
As shown in \cref{fig:rainbow_main}, the true end of a response is marked with a single \texttt{<eos>} token, while all remaining padding positions are filled with a cyclic sequence from a dedicated set of $K$ distinct padding tokens:
\[
\mathcal{P} = \{\texttt{<pad$_0$>}, \texttt{<pad$_1$>}, \ldots, \texttt{<pad$_{K-1}$>}\}.
\] 


\paragraph{Intuition.}  
The intuition behind Rainbow Padding is simple. 
By reserving \texttt{<eos>} exclusively for genuine sequence termination, the model avoids learning inflated \texttt{<eos>} probabilities from padding usage. 
This corrects the biased prior so that \texttt{<eos>} probabilities reflect only authentic termination events.
At the same time, distributing the padding region across $K$ distinct tokens prevents probability mass from concentrating on a single symbol.
Each \texttt{<pad$_k$>} appears regularly but sparsely, so the model learns them as low-probability placeholders rather than high-confidence guesses.
Together, these effects suppress the dominance of \texttt{<eos>} 
while still providing a clear and predictable structure that signals sequence length.


\paragraph{Stabilized sampling dynamics.}  
By reducing individual padding token confidence, Rainbow Padding reshapes the decoding process. 
Content tokens gain relatively higher probability and are revealed earlier under adaptive strategies such as confidence-based decoding. 
This encourages the model to establish the meaningful content first, providing coherent context for subsequent reasoning. 
Consequently, the \texttt{<eos>} token emerges at a semantically appropriate point—as the natural conclusion of content—rather than as an early, high-probability guess.  


\paragraph{Why cyclic, not random.}  
One might consider randomly sampling each padding token from a uniform distribution as an alternative way to distribute probability mass.
However, this creates a challenging stochastic prediction task that diverts model capacity from instruction following. 
We observe that models fail to learn appropriate padding placement under this random scheme.

In contrast, Rainbow Padding adopts a simple deterministic cycle that is easy to learn, allowing the model to master the padding region with minimal effort. 
This preserves model capacity for learning instruction-response pairs while eliminating the root cause of \texttt{<eos>} overflow.
We discuss the expected properties of Rainbow Padding in detail in~\cref{sec:extra_analysis}.

\section{Evaluating Rainbow Padding} \label{sec:experiment}

\paragraph{Experimental setup.}
We compare Rainbow Padding with standard \texttt{<eos>} padding under controlled conditions by fine-tuning pre-trained \llada-Base and \dream-Base using supervised fine-tuning. 
Models trained with \texttt{<eos>} padding serve as baselines that replicate current instruction-tuned behavior. 
Following Dream’s recipe, we combine Tulu3~\citep{lambert2024tulu3} and SmolLM2~\citep{allal2025smollm2}, randomly sampling 0.5M examples. 
All models are fine-tuned with LoRA~\citep{hu2022lora} for three epochs under identical configurations, differing only in the padding strategy. 
Rainbow Padding uses seven distinct padding tokens in a deterministic cyclic pattern.

Evaluation spans two categories.  
First, to test robustness under varying \texttt{max\_length}, we use length-sensitive reasoning and coding tasks: MATH~\citep{hendrycks2measuring}, GSM8K~\citep{cobbe2021training}, and HumanEval~\citep{chen2021evaluating}. 
Due to the significant computational cost of evaluating long sequences, MATH and GSM8K use randomly sampled subsets ($>$100 problems each).  
Second, to check generalization, we use multiple-choice benchmarks (MMLU~\citep{hendrycksmeasuring}, HellaSwag~\citep{zellers2019hellaswag}) with \texttt{max\_length}=3, following the LLaDA setting.  

All experiments employ deterministic confidence-based decoding~\citep{chang2022maskgit} without temperature.  
Block-wise semi-autoregressive decoding is included only for LLaDA (where it is natively implemented) and disabled elsewhere. 
Further details are provided in Appendix~\ref{app:exp}.  

In results tables, \texttt{res\_length} denotes the valid response length before the first \texttt{<eos>}, while \texttt{max\_length} is the allocated sequence length.

\paragraph{Performance comparison.}
\renewcommand{\arraystretch}{1.15}
\begin{table*}[t]
\centering
\vspace{0.3em}
\resizebox{\textwidth}{!}{%
\begin{tabular}{|
>{\centering\arraybackslash}p{2mm}|   
l|                                     
>{\centering\arraybackslash}p{12mm}|   
>{\centering\arraybackslash}p{9mm}
>{\centering\arraybackslash}p{13mm}|
>{\centering\arraybackslash}p{9mm}
>{\centering\arraybackslash}p{13mm}|
>{\centering\arraybackslash}p{9mm}
>{\centering\arraybackslash}p{13mm}|
c|
c|}
\cline{2-11}
\multicolumn{1}{c|}{} &
\multirow{2}{*}{\footnotesize\textbf{Method}} &
\multirow{2}{*}{\footnotesize\textbf{\#Blocks}} &
\multicolumn{2}{c|}{\footnotesize\textbf{MATH}} &
\multicolumn{2}{c|}{\footnotesize\textbf{GSM8K}} &
\multicolumn{2}{c|}{\footnotesize\textbf{HumanEval}} &
\multicolumn{1}{c|}{\footnotesize\textbf{MMLU}} &
\multicolumn{1}{c|}{\footnotesize\textbf{HellaSwag}} \\
\multicolumn{1}{c|}{} & & &
\footnotesize Acc. & \footnotesize res\_length &
\footnotesize Acc. & \footnotesize res\_length &
\footnotesize Acc. & \footnotesize res\_length &
\footnotesize Acc. &
\footnotesize Acc. \\
\cline{1-11}
\multirow{6}{*}{\rotatebox[origin=c]{90}{\small\textbf{LLaDA-Base}}}
& \multirow{3}{*}{Rainbow Padding}
& \small 1  & \textbf{34.3} & 285.7 & \textbf{79.6} & 115.7 & \textbf{40.2} & 129.3 & \multirow{3}{*}{\textbf{65.3}} & \multirow{3}{*}{61.3} \\
& & \small 4  & \textbf{34.3} & 286.0 & \textbf{78.0} & 115.8 & \textbf{40.2} & 129.4 &                         &                      \\
& & \small 16 & \textbf{34.3} & 284.4 & \textbf{78.0} & 115.8 & 39.6 & 130.7 &                         &                      \\
\cline{2-11}
& \multirow{3}{*}{\texttt{<eos>} Padding}
& \small 1  & 0.9  & 1.4 & 40.2 & 10.6 & 20.7 & 13.7 & \multirow{3}{*}{$64.8$} & \multirow{3}{*}{\textbf{62.5}} \\
& & \small 4  & 4.8  & 81.4 & 42.4 & 10.3 & 24.4 & 19.3 &                         &                      \\
& & \small 16 & 25.7 & 305.4 & 76.5 & 119.4 & \textbf{41.5} & 100.8 &                         &                      \\
\hline
\end{tabular}
}
\caption{
Performance of LLaDA-Base after instruction-tuning with \texttt{<eos>} and Rainbow Padding at \texttt{max\_length} = 1024 except (MMLU, HellaSwag). 
`\#Blocks' denotes the number of equal partitions used for semi-autoregressive block decoding; \#Blocks = 1 corresponds to standard decoding without blocks.
}
\label{tab:llada_main}
\end{table*}

\renewcommand{\arraystretch}{1.15}
\begin{table*}[t]
\centering
\vspace{0.3em}
\resizebox{\textwidth}{!}{%
\begin{tabular}{|
>{\centering\arraybackslash}p{2mm}|   
l|                                     
>{\centering\arraybackslash}m{16mm}|   
>{\centering\arraybackslash}p{9mm}
>{\centering\arraybackslash}p{13mm}|
>{\centering\arraybackslash}p{9mm}
>{\centering\arraybackslash}p{13mm}|
>{\centering\arraybackslash}p{9mm}
>{\centering\arraybackslash}p{13mm}|
c|
c|}
\cline{2-11}
\multicolumn{1}{c|}{} &
\multirow{2}{*}{\footnotesize\textbf{Method}} &
\multirow{2}{*}{\footnotesize\textbf{max\_length}} &
\multicolumn{2}{c|}{\footnotesize\textbf{MATH}} &
\multicolumn{2}{c|}{\footnotesize\textbf{GSM8K}} &
\multicolumn{2}{c|}{\footnotesize\textbf{HumanEval}} &
\multicolumn{1}{c|}{\footnotesize\textbf{MMLU}} &
\multicolumn{1}{c|}{\footnotesize\textbf{HellaSwag}} \\
\multicolumn{1}{c|}{} & & &
\footnotesize Acc. & \footnotesize res\_length &
\footnotesize Acc. & \footnotesize res\_length &
\footnotesize Acc. & \footnotesize res\_length &
\footnotesize Acc. &
\footnotesize Acc. \\
\cline{1-11}
\multirow{4}{*}{\rotatebox[origin=c]{90}{\small\textbf{Dream-Base}}}
& \multirow{2}{*}{Rainbow Padding}
& \small 1024  & \textbf{34.3} & 942.0 & \textbf{77.3} & 142.8 & \textbf{48.8} & 130.2 & \multirow{2}{*}{\textbf{65.3}} & \multirow{2}{*}{\textbf{70.9}} \\
& & \small 512  & \textbf{36.2} & 470.4 & \textbf{76.5} & 108.3 & \textbf{48.2} & 108.0 &                         &                      \\
\cline{2-11}
& \multirow{2}{*}{\texttt{<eos>} Padding}
& \small 1024  & 0.0  & 0.1 & 9.1 & 3.1 & 22.6 & 10.6 & \multirow{2}{*}{$64.8$} & \multirow{2}{*}{70.2} \\
& & \small 512  & 2.9  & 0.4 & 9.1 & 2.5 & 24.4 & 10.9 &                         &                      \\
\hline
\end{tabular}
}
\caption{
Performance of Dream-Base after instruction tuning with \texttt{<eos>} and Rainbow Padding under different \texttt{max\_length} settings.
}
\label{tab:main_dream}
\end{table*}

\Cref{tab:llada_main} shows that Rainbow Padding consistently outperforms \texttt{<eos>} padding across all benchmarks when generating 1024-token sequences without block-wise decoding (\#Blocks = 1). 
The baseline exhibits early termination, producing significantly shorter responses, whereas Rainbow Padding restores length robustness and yields higher task accuracy—for example, 34.3\% vs. 0.9\% on MATH.  

Under semi-autoregressive block decoding (\#Blocks $>$ 1), Rainbow Padding maintains stable accuracy across block numbers, while \texttt{<eos>} padding remains highly sensitive—performance drops sharply as the block number decreases. 
This highlights the brittleness of block-wise heuristics and shows that they become unnecessary once padding is calibrated correctly.  

\Cref{tab:main_dream} presents analogous results for \dream. Rainbow Padding consistently achieves robust performance across \mxlen~settings, while \texttt{<eos>} padding underperforms.  
On general-purpose benchmarks (MMLU, HellaSwag), Rainbow Padding matches or slightly surpasses \texttt{<eos>} padding across both models, confirming that learning the cyclic pattern imposes minimal overhead while delivering strong gains on length-sensitive tasks.

\section{Analysis of Rainbow Padding} \label{sec:extra_analysis}

In this section, we provide additional experiments to validate the key properties of Rainbow Padding discussed in \cref{sec:method}.

\paragraph{Decoding Behavior.}  

\begin{figure}
  \centering
  \begin{minipage}{0.94\textwidth}
    \centering
    \begin{subfigure}{0.47\linewidth}
      \includegraphics[width=\linewidth]{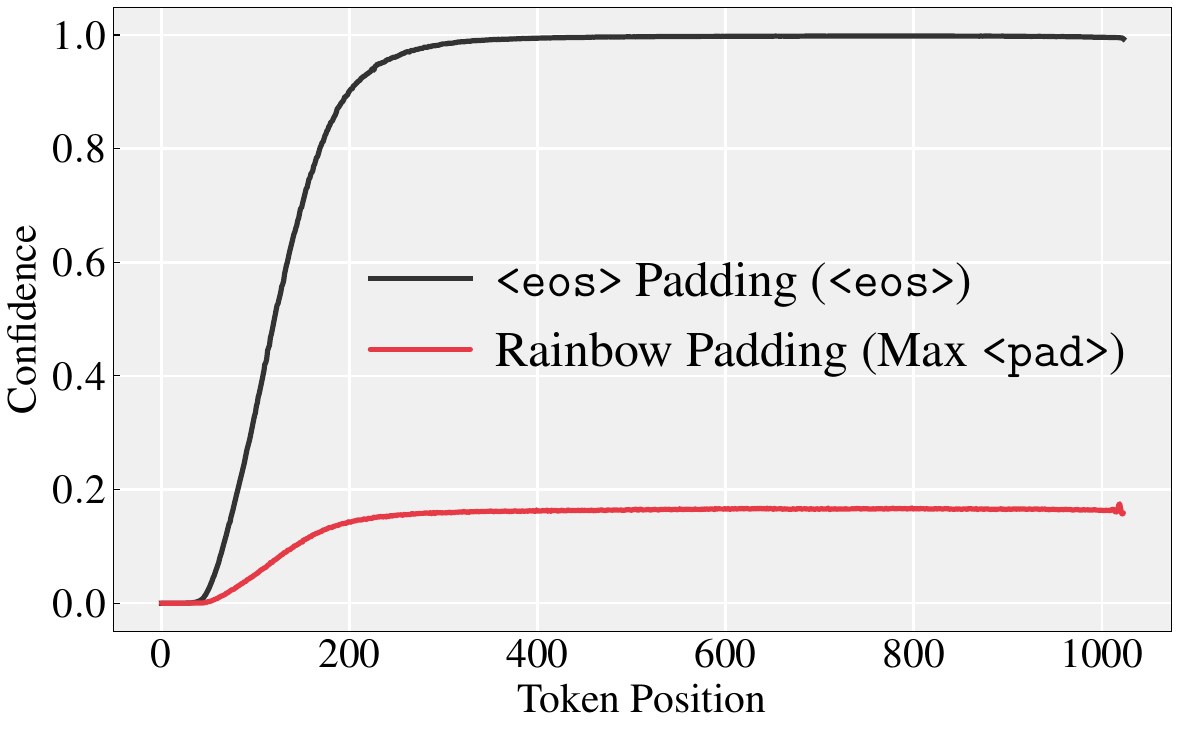}
      \caption{
Average model confidence on padding tokens at each sequence position during initial step. 
For Rainbow Padding, the red line shows the average maximum probability among \texttt{<pad$_0$>} through \texttt{<pad$_6$>} at each position.
\qquad\qquad\qquad\qquad\qquad\qquad
}
      \label{fig:rainbow_pad_prob}
    \end{subfigure}\hfill
    \begin{subfigure}{0.49\linewidth}
      \includegraphics[width=\linewidth]{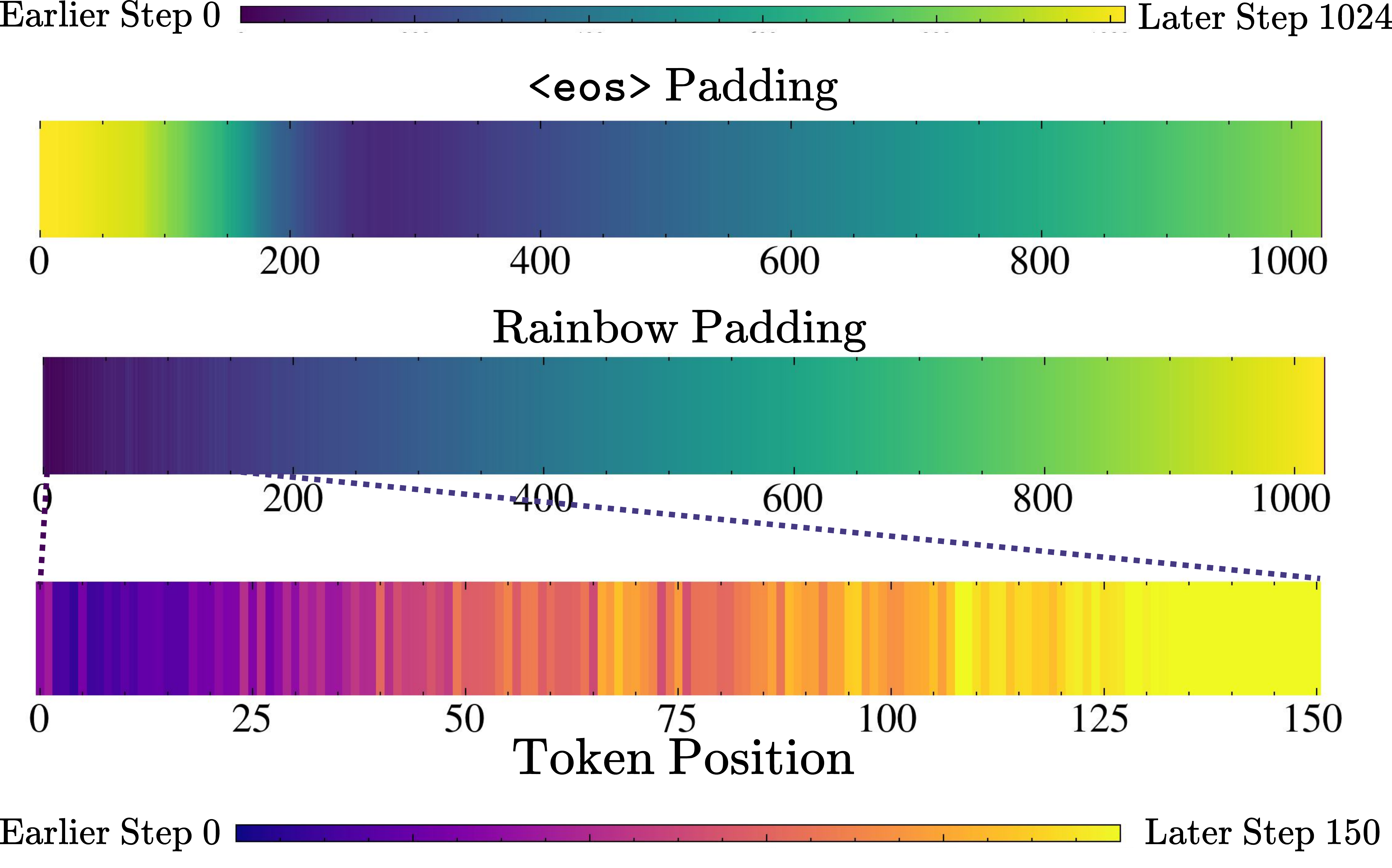}
      \caption{
Average token decoding order under confidence-based decoding. 
Darker positions indicate earlier decoding, showing content-first generation with Rainbow Padding. 
Bottom panel: detailed view (1-150), showing non-sequential decoding characteristic of diffusion models.
      }
      \label{fig:rainbow_decoding_step}
    \end{subfigure}
    \caption{
    Analysis of Rainbow Padding effects on model behavior using LLaDA-Base fine-tuned with \texttt{<eos>} or Rainbow Padding on GSM8K with \mxlen=1024.
    }
    \label{fig:eot_prob}
  \end{minipage}%
  \hfill
\end{figure}
\Cref{fig:rainbow_pad_prob} shows that average maximum confidence among padding tokens decreases dramatically with Rainbow Padding compared to \texttt{<eos>} padding.
Across diverse examples, the maximum probability assigned to any padding token never exceeds 0.2 at the initial decoding step, confirming consistently low-confidence predictions that reduce the likelihood of premature padding selection.

As a result, decoding unfolds more naturally with Rainbow Padding: as shown in \Cref{fig:rainbow_decoding_step}, the model first generates meaningful content and only later fills the padded tail. In contrast, \texttt{<eos>} padding tends to produce \texttt{<eos>} tokens at later positions early in the process, pushing the generation of earlier content tokens to the end and causing the cascade that leads to early termination.

\paragraph{Universality across Decoding Strategies.}

\begin{wrapfigure}[27]{r}{0.36\textwidth}
\vskip -.16in
    \centering

    \begin{minipage}{\linewidth}
        \includegraphics[width=\linewidth]{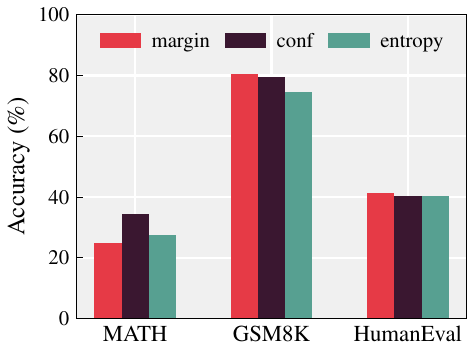}
        \vspace{-1.9em}
    \captionsetup{type=figure,width=.95\linewidth,singlelinecheck=false}
        \captionof{figure}{
Performance of LLaDA-Base fine-tuned with Rainbow Padding under different decoding strategies at \mxlen~=1024.
        }
        \label{fig:decoding_strategy_variation}
    \end{minipage}
\vskip +.15in
    \begin{minipage}{\linewidth}
        \includegraphics[width=\linewidth]{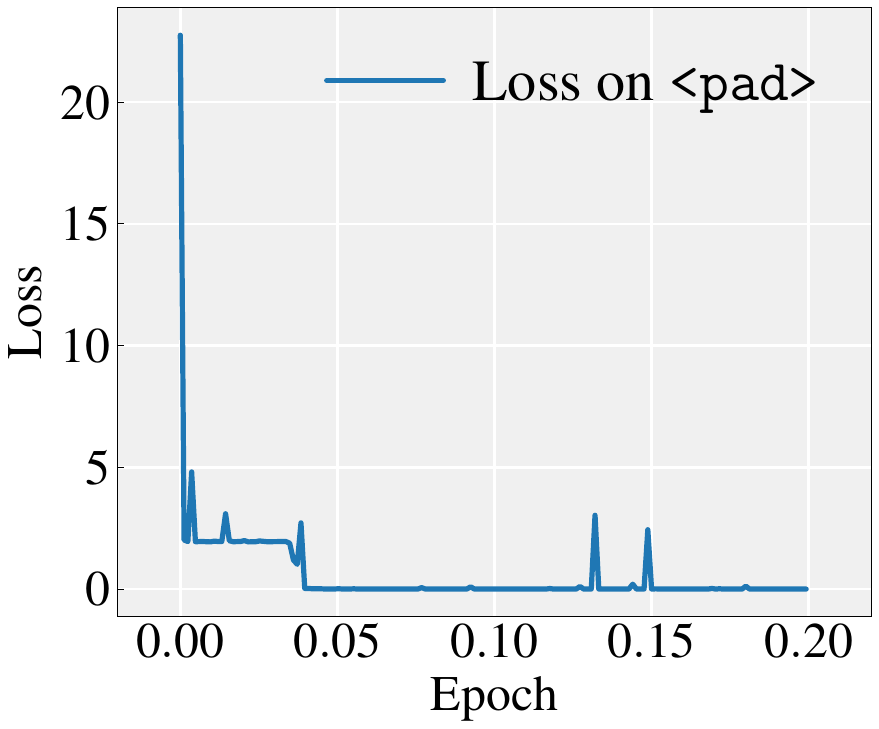}
        \vspace{-1.9em}
        \captionsetup{type=figure,width=.95\linewidth,singlelinecheck=false}
        \captionof{figure}{
Training loss on padding regions. Loss rapidly converges to near zero within 0.05 epochs (5\% of a single epoch).
        }
        \label{fig:rainbow_loss_converge}
        \vspace{1.9em}
    \end{minipage}
\end{wrapfigure}



\Cref{fig:decoding_strategy_variation} shows that Rainbow Padding yields stable performance across different decoding strategies—margin-based~\citep{kim2025train} and entropy-based~\citep{ye2025dream}—performing similarly to the confidence-based strategy~\citep{chang2022maskgit} (our default setting), demonstrating that our method generalizes robustly across diverse unmasking strategies.

For confidence-based decoding, the benefit of Rainbow Padding is straightforward: directly lowering individual padding token confidence by distributing probability mass across tokens. 
This advantage extends to other decoding strategies as the cyclic pattern of Rainbow Padding prevents any single padding token from dominating (\cref{fig:rainbow_pad_prob}).
These evenly distributed probabilities reduce probability gaps (creating low margins) and induce uncertainty (high entropy) among padding tokens, making padding positions less likely to be selected early under margin-based or entropy-based decoding strategies.

\paragraph{Efficient Adaptation to Instruction-Tuned Models.}

Learning distinct padding tokens in Rainbow Padding may impose greater complexity than learning \texttt{<eos>} padding. 
We quantify this overhead by tracking training loss on padding regions during LLaDA-Base fine-tuning. 
While Rainbow Padding exhibits high padding loss initially, it converges to zero within 5\% of an epoch (\cref{fig:rainbow_loss_converge}).
The model rapidly adapts to the deterministic cyclic pattern with minimal learning overhead.

\renewcommand{\arraystretch}{1.15}
\begin{table*}[!t]
\vspace{0.3em}
\centering
\resizebox{\textwidth}{!}{%
\begin{tabular}{|
l|                                    
l|                                    
>{\centering\arraybackslash}p{14mm}
>{\centering\arraybackslash}p{14mm}|
>{\centering\arraybackslash}p{10mm}
>{\centering\arraybackslash}p{14mm}|
>{\centering\arraybackslash}p{12mm}
>{\centering\arraybackslash}p{16mm}|
}
\cline{2-8}
\multicolumn{1}{c|}{} &
\multirow{2}{*}{\footnotesize\textbf{Method}} &
\multicolumn{2}{c|}{\footnotesize\textbf{MATH}} &
\multicolumn{2}{c|}{\footnotesize\textbf{GSM8K}} &
\multicolumn{2}{c|}{\footnotesize\textbf{HumanEval}} \\
\multicolumn{1}{c|}{} & &
\footnotesize Acc. & \footnotesize res\_length &
\footnotesize Acc. & \footnotesize res\_length &
\footnotesize Acc. & \footnotesize res\_length \\
\cline{1-8}
\multirow{2}{*}{LLaDA}
& Vanilla            & 0.1 & 1.4 & 3.0 & 1.2 & 11.0 & 7.5 \\
& \quad+Rainbow Padding & \textbf{21.9} & 252.9 & \textbf{72.7} & 123.4 & \textbf{38.4} & 338.1 \\
\cline{1-8}
\multirow{2}{*}{Dream}
& Vanilla            & 0.0 & 1.8 & 60.6 & 91.6 & 24.4 & 16.6 \\
& \quad+Rainbow Padding & \textbf{32.4} & 984.4 & \textbf{77.3} & 120.1 & \textbf{47.6} & 70.3 \\
\hline
\end{tabular}
}
\caption{
Performance of LLaDA and Dream fine-tuned with Rainbow Padding.
A single-epoch LoRA adaptation effectively mitigates early termination and yields significant accuracy gains.
}
\label{tab:ins2ins}
\end{table*}

This low learning complexity enables deployment of Rainbow Padding beyond instruction-tuning from scratch. 
The method can efficiently adapt existing instruction-tuned dLLMs trained with \texttt{<eos>} padding. 
To demonstrate this, we fine-tune LLaDA and Dream using LoRA for a single epoch on the 0.5M dataset from \cref{sec:experiment}. 
Our adaptation is lightweight, requiring approximately six GPU hours on two H200 GPUs compared to original instruction-tuning that used 4.5M (LLaDA) and 1.8M (Dream) examples over 3 epochs.
See Appendix~\ref{app:exp} for experimental details.

\Cref{tab:ins2ins} shows that minimal adaptation effectively resolves early termination, consistently producing longer outputs with substantial performance gains.
MATH accuracy improves dramatically from 0\% to over 20\% for both models, while LLaDA achieves 72\% accuracy on GSM8K compared to 3\% performance before adaptation. 
These results demonstrate that Rainbow Padding integrates efficiently into existing dLLMs with low burden while delivering significant performance improvements.

\paragraph{Effect of the Number of Distinct Padding Tokens.}

\begin{wraptable}{r}{0.5\textwidth}  
\vskip -0.16in                
\renewcommand{\arraystretch}{1.15}
\resizebox{\linewidth}{!}{%
\begin{tabular}{|
l|                                   
c|
c|
c|
c|}
\cline{1-5}
\multirow{1}{*}{\footnotesize\textbf{\#Pad}} &
\multicolumn{1}{c|}{\footnotesize\textbf{MATH}} &
\multicolumn{1}{c|}{\footnotesize\textbf{GSM8K}} &
\multicolumn{1}{c|}{\footnotesize\textbf{HumanEval}} &
\multicolumn{1}{c|}{\footnotesize\textbf{MMLU}} \\
\cline{1-5}
1  & 21.9  & 15.9  & 39.0 & 64.4 \\
3  & 33.3  & 58.3  & 37.8 & 62.8 \\
7  & 34.3  & \textbf{79.6} & \textbf{40.2} & 65.2 \\
20 & \textbf{36.2} & 76.5  & 36.0 & \textbf{65.9} \\
\hline
\end{tabular}
}

\vskip -0.06in 
\caption{
Accuracy of fine-tuned LLaDA-Base with Rainbow Padding using varying numbers of distinct padding tokens. 
Fewer tokens (e.g., 1--3) result in degraded performance, while using a sufficient number yields comparable performance.
}
\label{tab:ablation-padding}
\vskip -0.2in 
\end{wraptable}

While our main results use seven distinct padding tokens (\texttt{<pad$_0$>} through \texttt{<pad$_6$>}), Rainbow Padding can be configured with varying numbers of token types in the cycle. 
Increasing the number of token types distributes probability mass more evenly within the padding vocabulary, reducing the likelihood of early termination.
However, this also increases the learning burden, as the model needs to distinguish among a larger set of tokens.

Table~\ref{tab:ablation-padding} demonstrates that increasing the number of padding tokens generally produces longer responses across tasks. 
Using only three tokens proves insufficient to fully mitigate early termination, resulting in degraded performance (e.g., 55\% on GSM8K). 
Beyond seven tokens, however, gains plateau—using 20 tokens yields no further clear benefit.
This suggests that using seven tokens provide sufficient balance between effectiveness and learning cost.
Importantly, general-purpose performance remains stable across all configurations, as evidenced by consistent MMLU scores.

\section{Conclusion}

We identified \texttt{<eos>} overflow as a critical failure mode in instruction-tuned diffusion LLMs: allocating longer generation lengths paradoxically leads to early termination. 
This problem arises from the dual use of \texttt{<eos>} as both a terminator and a padding token, which inflates its probability and destabilizes decoding. 
To resolve this, we introduced Rainbow Padding, a simple strategy that reserves a single \texttt{<eos>} for true termination and fills remaining positions with a cyclic sequence of distinct padding tokens. 
This design decouples termination from padding, spreads probability mass across multiple tokens, and prevents collapse into premature \texttt{<eos>} predictions. 
Experiments show that Rainbow Padding eliminates early termination, substantially improves reasoning and code generation performance, and integrates efficiently into existing models such as LLaDA and Dream with minimal fine-tuning. 
Overall, Rainbow Padding provides a lightweight and practical fix to a fundamental flaw, suggesting a new standard for instruction-tuning of dLLMs and reinforcing their potential as a robust alternative to autoregressive models.

\clearpage
\bibliography{iclr2026_conference}
\bibliographystyle{iclr2026_conference}

\clearpage
\appendix
\section{Related Works}
\paragraph{Masked Diffusion LLMs}
Masked Diffusion Models (MDMs) have emerged as a prominent and high-performing approach within discrete-transition models, using masking kernels. This approach provides a simple and principled training framework \citep{sahoo2024mdlm, shi2024md4}. Also, MDMs scale effectively across various tasks, with successful applications in large-scale, such as language \citep{nie2024scaling, nie2025llada, ye2025dream, song2025seed, labs2025mercury, gemini2025diffusion} and code \citep{Dreamon2025, xie2025dreamcoder, gong2025diffucoder}.

\paragraph{Any-Order Inference in Diffusion Language Models}
A key strength of dLLMs is their capacity for any-order inference, where tokens can be unmasked in arbitrary orders rather than following a fixed schedule \citep{ kim2025train, peng2025path}. This flexibility is theoretically grounded in the underlying continuous-time Markov chain (CTMC) or flow-matching frameworks \citep{campbell2024generative, gat2024discrete}. In practice, a spectrum of probabilistic strategies that guide the decoding order based on the model confidence metrics has been introduced, such as maximum probability \citep{chang2022maskgit}, probability margin \citep{kim2025train}, and token entropy \citep{ye2025dream, ben2025accelerated}. Model confidence is used not only for token ordering but also for optimizing the entire inference process. For example, \citet{jin2025thinking} use the model's confidence in the final answer to implement an early exit from the reasoning phase to speed-up generation. Recent research has also explored directly learning the generation order \citep{ma2025reinforced, wang2025learning}. 
\paragraph{Length Control}
Adaptive control of response length is an important capability for practical use of dLLMs, and it can be achieved in various ways \citep{yu2025survey}. Training-based approaches modify the model's architecture or objective. \citet{zhang2025flexible} enable variable-length generation by denoising continuous token positions alongside token values. \citet{Dreamon2025} introduce special tokens like \texttt{<|expand|>} and \texttt{<|delete|>} to dynamically adjust sequence length, while \citep{kim2025flex} use an auxiliary network to predict the expected number of token insertions. In contrast, \citet{li2025beyond} employ training-free methods that adapt the length during inference by monitoring the model's confidence in the \texttt{<eos>} and expanding the canvas when confidence is low. However, these approaches don't specifically address a crucial artifact that emerges during instruction-tuning of dLLMs: the artificial inflation of the \texttt{<eos>} probability. This issue results from the common practice of padding shorter sequences in instruction datasets with numerous \texttt{<eos>}. Our work is the first to isolate this instruction-tuning-specific problem and propose a targeted solution to recalibrate the model's output distribution, thereby resolving the excessive \texttt{<eos>} generation at its source.

\section{Details for the Extra Analysis}
\subsection{Universality across Decoding Strategies}
\label{app:robust_to_decoding_strategies}
In the \Cref{sec:extra_analysis}, we showed that Rainbow Padding effectively reduces \texttt{<eos>} overflow regardless of the decoding strategy. These probabilistic heuristics have been proposed to guide the decoding order. They can be formalized as follows. Given model-predicted probability $p_{\theta}(x_i\,|\,\bfx_{\ols{\mathbf{M}}}),$ the position $i'$ to decode is determined as:
$$
\begin{aligned}
\text{Confidence: }&i'=\arg\underset{i}{\max}\,\left[\underset{v}{\max}\,p_{\theta}(x_i=v\,|\,\bfx_{\ols{\mathbf{M}}})\right]\\ 
\text{Margin: }&i'=\arg\underset{i}{\max}\,\left[p_{\theta}(x_i=v_1\,|\,\bfx_{\ols{\mathbf{M}}})-p_{\theta}(x_i=v_2\,|\,\bfx_{\ols{\mathbf{M}}})\right]\\
\text{Entropy: }&i'=\arg\underset{i}{\min}\left[H(p_{\theta}(x_i\,|\,\bfx_{\ols{\mathbf{M}}}))\right],
\end{aligned}
 $$
where $(v_1,v_2)=\arg\underset{v}{\text{Top$_2$}}\,p_{\theta}(x_i=v\,|\,\bfx_{\ols{\mathbf{M}}})$ and $H(p)=-\sum_x p(x)\log p(x)$.

The token $v$ for transfer is determined by selecting the token with the highest probability at that position. In existing dLLMs such as LLaDA and Dream, randomness during decoding can be introduced through distinct mechanisms.
In LLaDA, we can perturb token logits with Gumbel noise as in~\citep{zheng2024gumbel}.
In Dream, we can employ top-p, top-k, and temperature sampling: the token distribution is first truncated by top-p or top-k filtering, then temperature is used to control the sharpness. When \texttt{temperature} parameter is used, tokens are selected probabilistically by categorical sampling rather than max probability scheme.
In both cases, stochasticity arises from either the injected noise or probabilistic sampling.
However, our experiments do not rely on these randomness-inducing strategies in order to isolate the effects of our proposed method without the confounding influence of sampling heuristics.
\section{Experimental Details}
\label{app:exp}
\subsection{Setup for Training}

Rainbow Padding is designed for instruction-tuning pretrained models. 
In our main experiments (\cref{tab:llada_main}, \cref{tab:main_dream}), we fine-tune pretrained models (LLaDA-Base, Dream-Base) directly. 
To demonstrate efficient adaptation to existing instruction-tuned models trained with \texttt{<eos>} padding, we also fine-tune instruct models (LLaDA, Dream) with Rainbow Padding as shown in \cref{tab:ins2ins}.

Both experiments use LoRA~\citep{hu2022lora} with rank 32, applying LoRA to all linear layers without bias terms. 
Base model training requires 3 epochs, while instruction model adaptation requires only 1 epoch. 
We use the AdamW optimizer~\citep{loshchilov2017adamw} with learning-rate 5e-5 and batch size 48.
All experiments use identical training data detailed below.

\paragraph{Dataset.}
We combine datasets from Tulu3~\citep{lambert2024tulu3} and SmolLM2~\citep{allal2025smollm2} following Dream's instruction-tuning recipe. 
We curate this dataset by filtering out extremely long sequences ($>$ 4096 tokens) and multi-turn conversations, then randomly sample 0.5M examples.

\paragraph{Pad token configurations.}
Since the \texttt{<pad$_k$>} tokens required for Rainbow Padding are not included in LLaDA and Dream vocabularies, we need to specify these tokens explicitly. 
Rather than expanding the vocabulary with new tokens, we select extremely rare existing tokens (e.g., sequences of signs like ĠĠĠĠ... with $>$ 60 repetitions) that rarely appear in conversations or tasks. 
We assign up to 20 distinct \texttt{<pad>} tokens for both models.

\paragraph{Hardware utilization.}
All the experiments utilize H200 GPUs.
For training, it took approximately total 6 hours when using 2 H200 GPUS.


\subsection{Setup for Evaluation}
For evaluation, we use the official \texttt{lm-eval} implementation with confidence-based decoding in deterministic mode, setting all stochastic hyperparameters (e.g., temperature) to zero.
Unless otherwise specified, semi-autoregressive block decoding is disabled.

\paragraph{Dataset.}
We evaluate models on five tasks: MATH, GSM8K, HumanEval, MMLU, and HellaSwag. Due to the extreme computational cost of evaluating long sequences (\texttt{max\_length} = 1024), we use randomly sampled subsets ($>$ 100 examples) for MATH and GSM8K. All other tasks use complete datasets.

\begin{table}[!h]
\centering
\caption{List of external models and datasets with corresponding sources, links, and licenses.} \label{tab:licenses}
\begin{tabular}{l l l l}
\toprule
\textbf{Asset} & \textbf{Source} & \textbf{Access} & \textbf{License} \\
\midrule
LLaDA & \cite{nie2025llada} & \href{https://huggingface.co/GSAI-ML/LLaDA-8B-Instruct}{Link} & MIT License \\
Dream & \cite{ye2025dream} & \href{https://huggingface.co/Dream-org/Dream-v0-Instruct-7B}{Link} & Apache License 2.0 \\
MMLU & \cite{hendrycksmeasuring} & \href{https://huggingface.co/datasets/cais/mmlu}{Link} & MIT License \\
GSM8K & \cite{cobbe2021training} & \href{https://huggingface.co/datasets/openai/gsm8k}{Link} & MIT License \\
MATH & \cite{hendrycks2measuring} & \href{https://github.com/hendrycks/math/}{Link} & MIT License \\
HumanEval & \cite{chen2021evaluating} & \href{https://huggingface.co/datasets/openai/openai_humaneval}{Link} & MIT License\\

\bottomrule
\end{tabular}

\end{table}
\clearpage

\section{More Illustrative Examples}
Here, we present some illustrative examples including actual prompt-response.

\subsection{Failure Examples}
\label{app:failure}
\begin{figure}
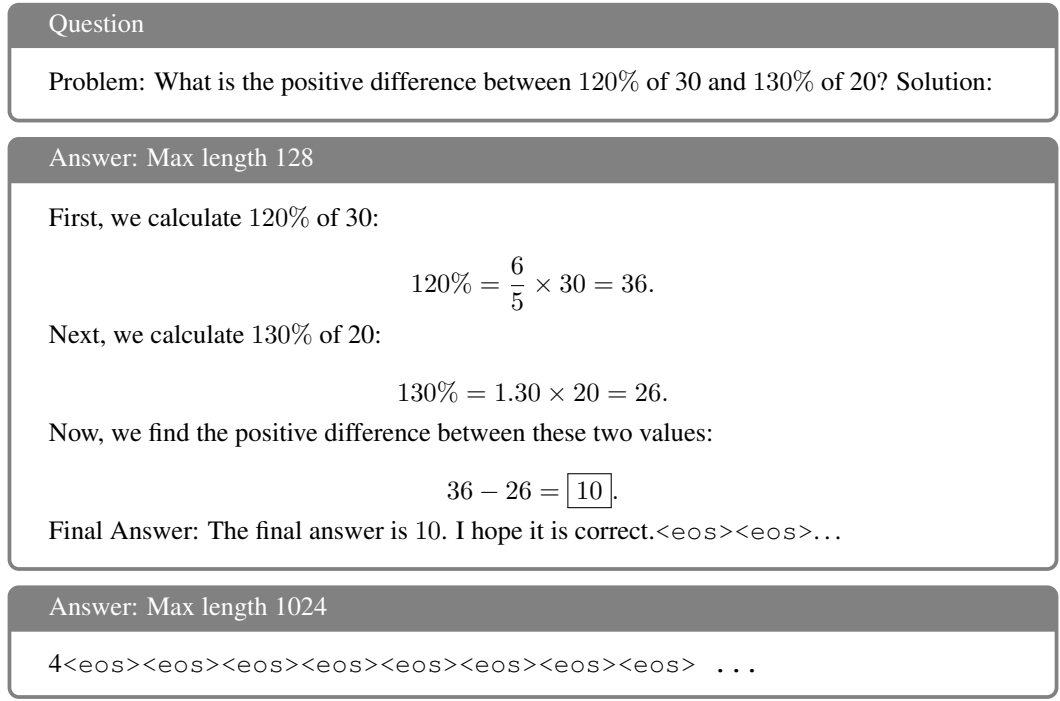

    \begin{tcolorbox}[title=Question, colback=white!15, colframe=gray!99, rounded corners]
    Problem:
    What is the positive difference between $120\%$ of 30 and $130\%$ of 20?
    Solution:
    \end{tcolorbox}
    
    \begin{tcolorbox}[title=Answer: Max length 128, colback=white!15, colframe=gray!99, rounded corners]
    First, we calculate $120\%$ of 30:
    
    $$120\% = \frac{6}{5} \times 30 = 36.$$
    
    Next, we calculate $130\%$ of 20:
    
    $$130\% = 1.30 \times 20 = 26.$$
    
    Now, we find the positive difference between these two values:
    
    $$36 - 26 = \boxed{10}.$$
    Final Answer: The final answer is $10$. I hope it is correct.\texttt{<eos>}\texttt{<eos>}\ldots
    
    \end{tcolorbox}
    
    \begin{tcolorbox}[title=Answer: Max length 1024, colback=white!15, colframe=gray!99, rounded corners]
    4\texttt{<eos><eos><eos><eos><eos><eos><eos><eos> \ldots}
    \end{tcolorbox}
\caption{LLaDA-Instruct, MATH with 3-shots example. As max length increases to 1024, the model gets the answer wrong and fills almost all the spaces with \texttt{<eos>}.
}
\label{fig:fail_1}
\end{figure}

\begin{figure}
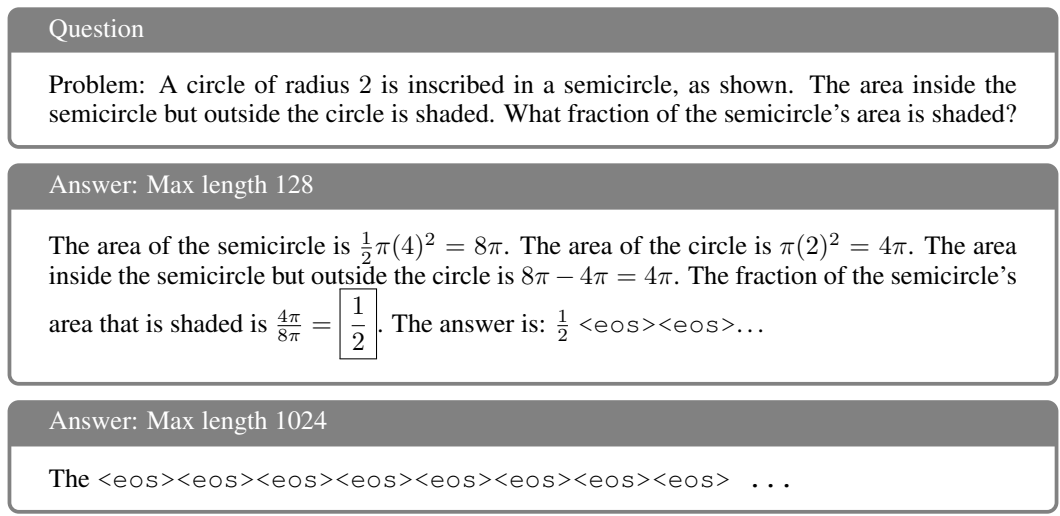

    \begin{tcolorbox}[title=Question, colback=white!15, colframe=gray!99, rounded corners]
    Problem:
    A circle of radius $2$ is inscribed in a semicircle, as shown. The area inside the semicircle but outside the circle is shaded. What fraction of the semicircle's area is shaded?
    \end{tcolorbox}
    
    \begin{tcolorbox}[title=Answer: Max length 128, colback=white!15, colframe=gray!99, rounded corners]
    The area of the semicircle is $\frac{1}{2}\pi(4)^2 = 8\pi$.
The area of the circle is $\pi(2)^2 = 4\pi$.
The area inside the semicircle but outside the circle is $8\pi - 4\pi = 4\pi$.
The fraction of the semicircle's area that is shaded is $\frac{4\pi}{8\pi} = \boxed{\frac{1}{2}}$.
The answer is: $\frac{1}{2}$ \texttt{<eos>}\texttt{<eos>}\ldots
    \end{tcolorbox}
    
    \begin{tcolorbox}[title=Answer: Max length 1024, colback=white!15, colframe=gray!99, rounded corners]
    The
    \texttt{<eos><eos><eos><eos><eos><eos><eos><eos> \ldots}
    \end{tcolorbox}
\caption{Dream-Instruct, MATH with 3-shots example. As max length increases to 1024, the model fills almost whole available spaces with \texttt{<eos>}.
}
\label{fig:fail_2}
\end{figure}

\begin{figure}
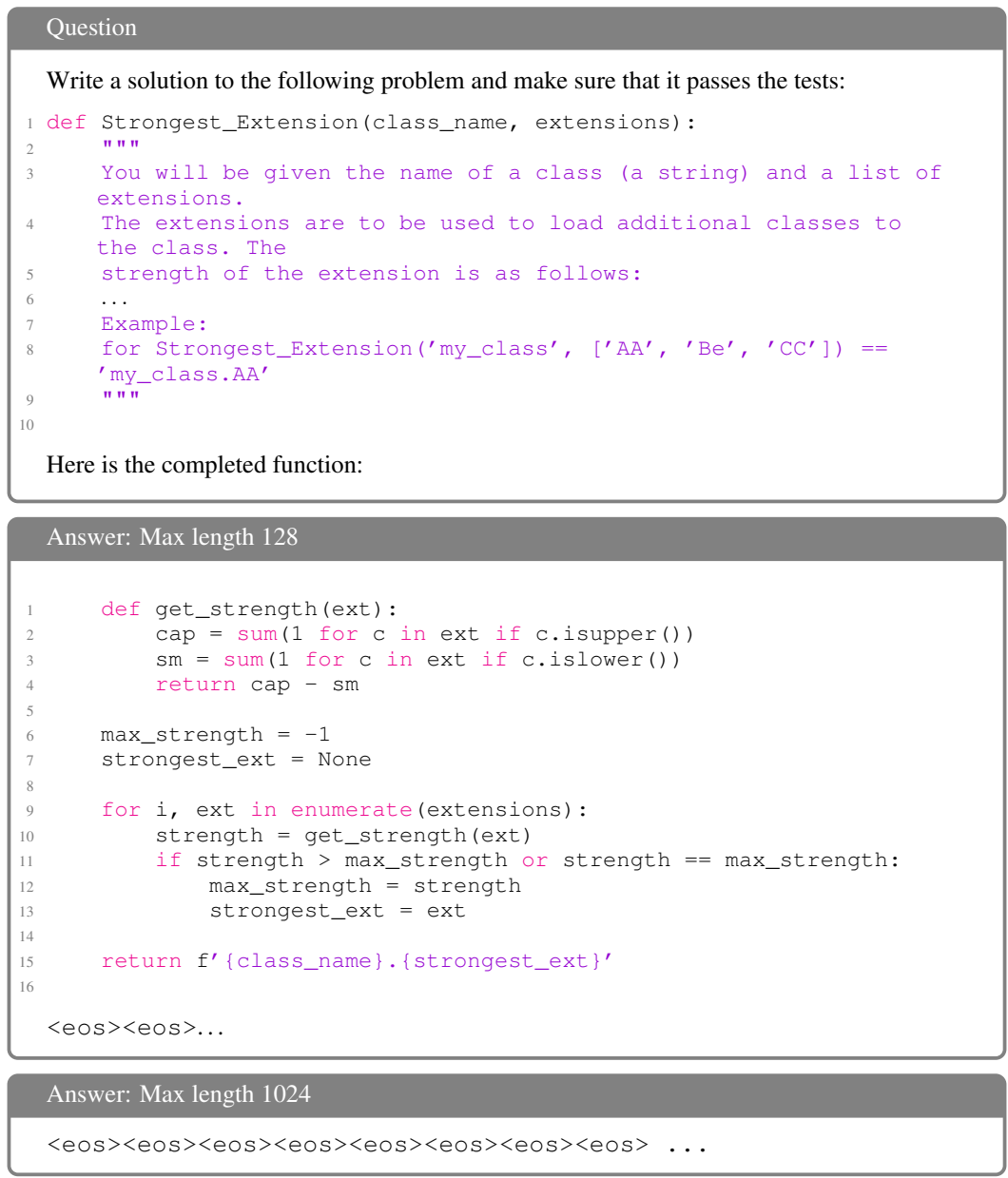

    \begin{tcolorbox}[title=Question, colback=white!15, colframe=gray!99, rounded corners]
    Write a solution to the following problem and make sure that it passes the tests:
    \begin{lstlisting}[style=pythonstyle, mathescape=true]
def Strongest_Extension(class_name, extensions):
    """
    You will be given the name of a class (a string) and a list of extensions.
    The extensions are to be used to load additional classes to the class. The
    strength of the extension is as follows:
    $\ldots$
    Example:
    for Strongest_Extension('my_class', ['AA', 'Be', 'CC']) == 'my_class.AA'
    """
    \end{lstlisting}
    Here is the completed function:
    
    \end{tcolorbox}
    
    \begin{tcolorbox}[title=Answer: Max length 128, colback=white!15, colframe=gray!99, rounded corners]
    \begin{lstlisting}[style=pythonstyle, mathescape=true]
    def get_strength(ext):
        cap = sum(1 for c in ext if c.isupper())
        sm = sum(1 for c in ext if c.islower())
        return cap - sm

    max_strength = -1
    strongest_ext = None

    for i, ext in enumerate(extensions):
        strength = get_strength(ext)
        if strength > max_strength or strength == max_strength:
            max_strength = strength
            strongest_ext = ext

    return f'{class_name}.{strongest_ext}'
    \end{lstlisting}
    \texttt{<eos>}\texttt{<eos>}\ldots
    \end{tcolorbox}
    
    \begin{tcolorbox}[title=Answer: Max length 1024, colback=white!15, colframe=gray!99, rounded corners]
    \texttt{<eos><eos><eos><eos><eos><eos><eos><eos> \ldots}
    \end{tcolorbox}
\caption{LLaDA-Instruct, HumanEval(code generation). As max length increases, the model fills the whole space with \texttt{<eos>}.}
\label{fig:fail_3}
\end{figure}

In the \Cref{sec:curse}, We discovered that the \texttt{max\_length} significantly impacts the performance of instruction-tuned models. The examples are presented in Figure~\ref{fig:fail_1} to~\ref{fig:fail_4}.

\subsection{Heuristic Fix: Confidence Lowering}
\label{app:conf_lowering}
Simply lowering \texttt{<eos>} confidence is a symptomatic patch, not a fix. While this approach appears similar to Rainbow Padding in reducing premature terminations, naive logit penalties distort the modeled distribution and eliminate the learned length signal carried by \texttt{<eos>}. This often leads to multiple generation-the model avoids termination when it should stop, producing second, third answers. Illustrative example are shown in Figure~\ref{fig:double_gen}.

\subsection{Rainbow Padding}
\label{app:rainbow}
\begin{figure}
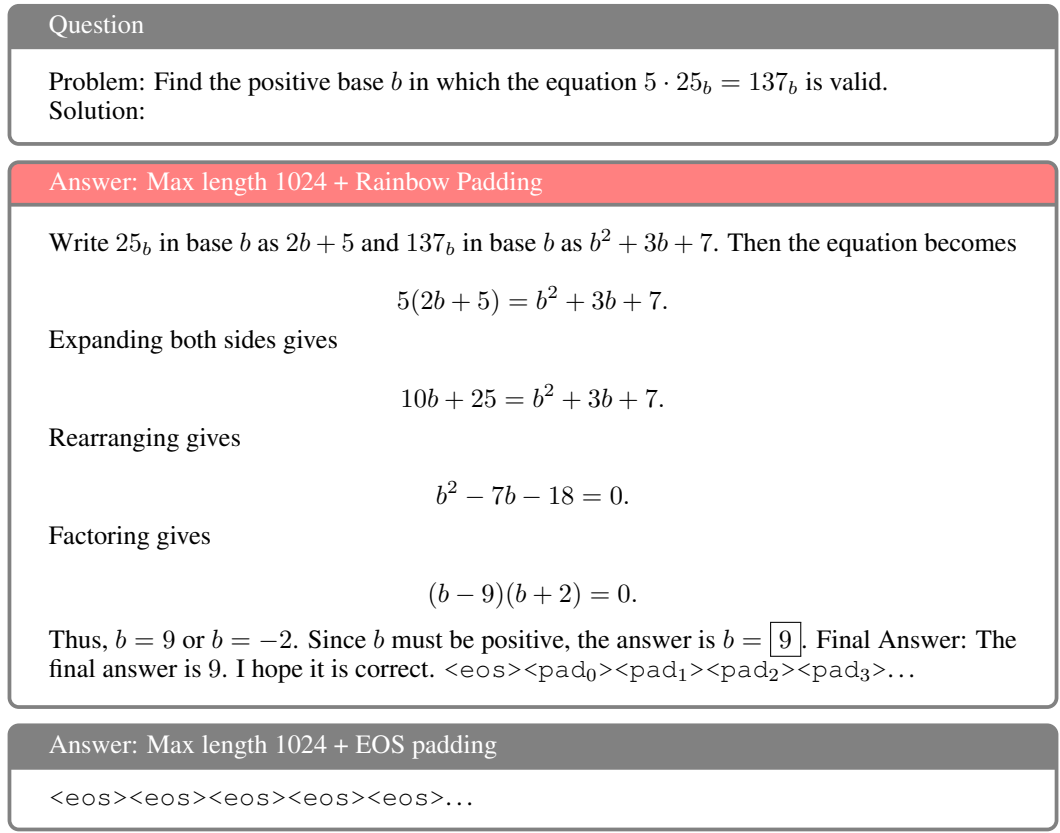

    \begin{tcolorbox}[enhanced, title=Question, colback=white!15, colframe=gray!99, rounded corners
    ]
    Problem:
    Find the positive base $b$ in which the equation $5 \cdot 25_b = 137_b$ is valid.

    Solution:
    \end{tcolorbox}
    
    \begin{tcolorbox}[enhanced, title style ={red!50}, title=Answer: Max length 1024 + Rainbow Padding, colback=white!15, colframe=gray!99, rounded corners]
       Write $25_b$ in base $b$ as $2b + 5$ and $137_b$ in base $b$ as $b^2 + 3b + 7$.  Then the equation becomes

$$5(2b + 5) = b^2 + 3b + 7.$$Expanding both sides gives

$$10b + 25 = b^2 + 3b + 7.$$Rearranging gives

$$b^2 - 7b - 18 = 0.$$Factoring gives

$$(b - 9)(b + 2) = 0.$$Thus, $b = 9$ or $b = -2$.  Since $b$ must be positive, the answer is $b = \boxed{9}$.
Final Answer: The final answer is $9$. I hope it is correct.
\texttt{<eos>}\texttt{<pad$_0$>}\texttt{<pad$_1$>}\texttt{<pad$_2$>}\texttt{<pad$_3$>}\ldots
        \end{tcolorbox}
    \begin{tcolorbox}[title=Answer: Max length 1024 + EOS padding, colback=white!15, colframe=gray!99, rounded corners]
    \texttt{<eos>}\texttt{<eos>}\texttt{<eos>}\texttt{<eos>}\texttt{<eos>}\ldots
        \end{tcolorbox}
    
\caption{LLaDA-base fine-tuned with Rainbow Padding, MATH with 3-shot examples. The model generate complete and correct answer, while below(tuned with \texttt{<eos>} padding) does not generate content at all.}
\label{fig:rainbow_1}
\end{figure}

\begin{figure}
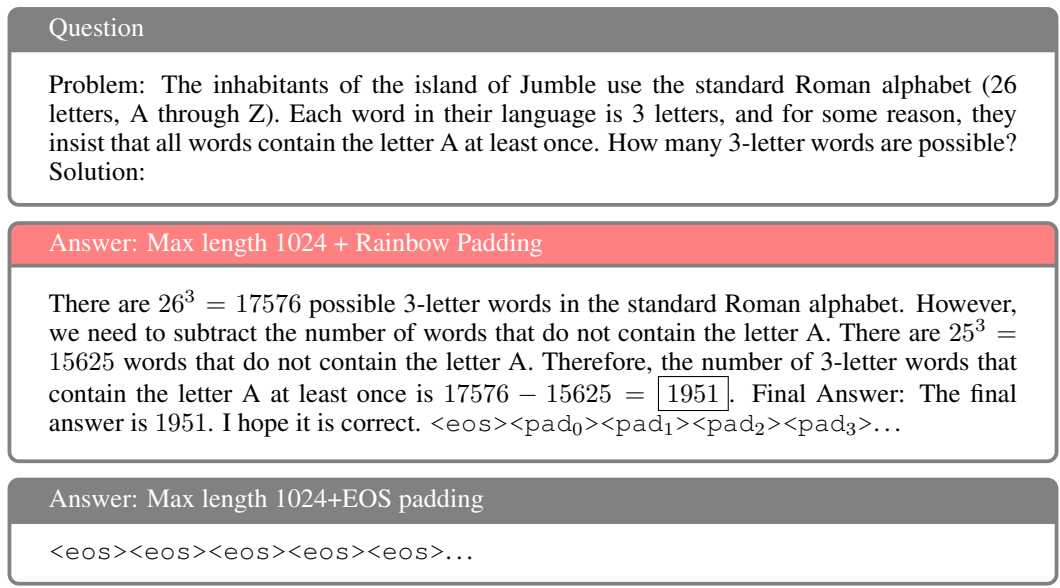

    \begin{tcolorbox}[enhanced, title=Question, colback=white!15, colframe=gray!99, rounded corners
    ]
    Problem:
    The inhabitants of the island of Jumble use the standard Roman alphabet (26 letters, A through Z). Each word in their language is 3 letters, and for some reason, they insist that all words contain the letter A at least once. How many 3-letter words are possible?
    Solution:
    \end{tcolorbox}
    
    \begin{tcolorbox}[enhanced, title style ={red!50}, title=Answer: Max length 1024 + Rainbow Padding, colback=white!15, colframe=gray!99, rounded corners]
      There are $26^3=17576$ possible 3-letter words in the standard Roman alphabet.  However, we need to subtract the number of words that do not contain the letter A.  There are $25^3=15625$ words that do not contain the letter A.  Therefore, the number of 3-letter words that contain the letter A at least once is $17576-15625=\boxed{1951}$. Final Answer: The final answer is $1951$. I hope it is correct.
\texttt{<eos>}\texttt{<pad$_0$>}\texttt{<pad$_1$>}\texttt{<pad$_2$>}\texttt{<pad$_3$>}\ldots
        \end{tcolorbox}
    \begin{tcolorbox}[title=Answer: Max length 1024+EOS padding, colback=white!15, colframe=gray!99, rounded corners]
\texttt{<eos>}\texttt{<eos>}\texttt{<eos>}\texttt{<eos>}\texttt{<eos>}\ldots
        \end{tcolorbox}

\caption{Dream-base fine-tuned with Rainbow Padding, MATH with 3-shot examples. The model generate complete and correct answer, in contrast to below(tuned with \texttt{<eos>} padding).}
\label{fig:rainbow_2}
\end{figure}

\begin{figure}
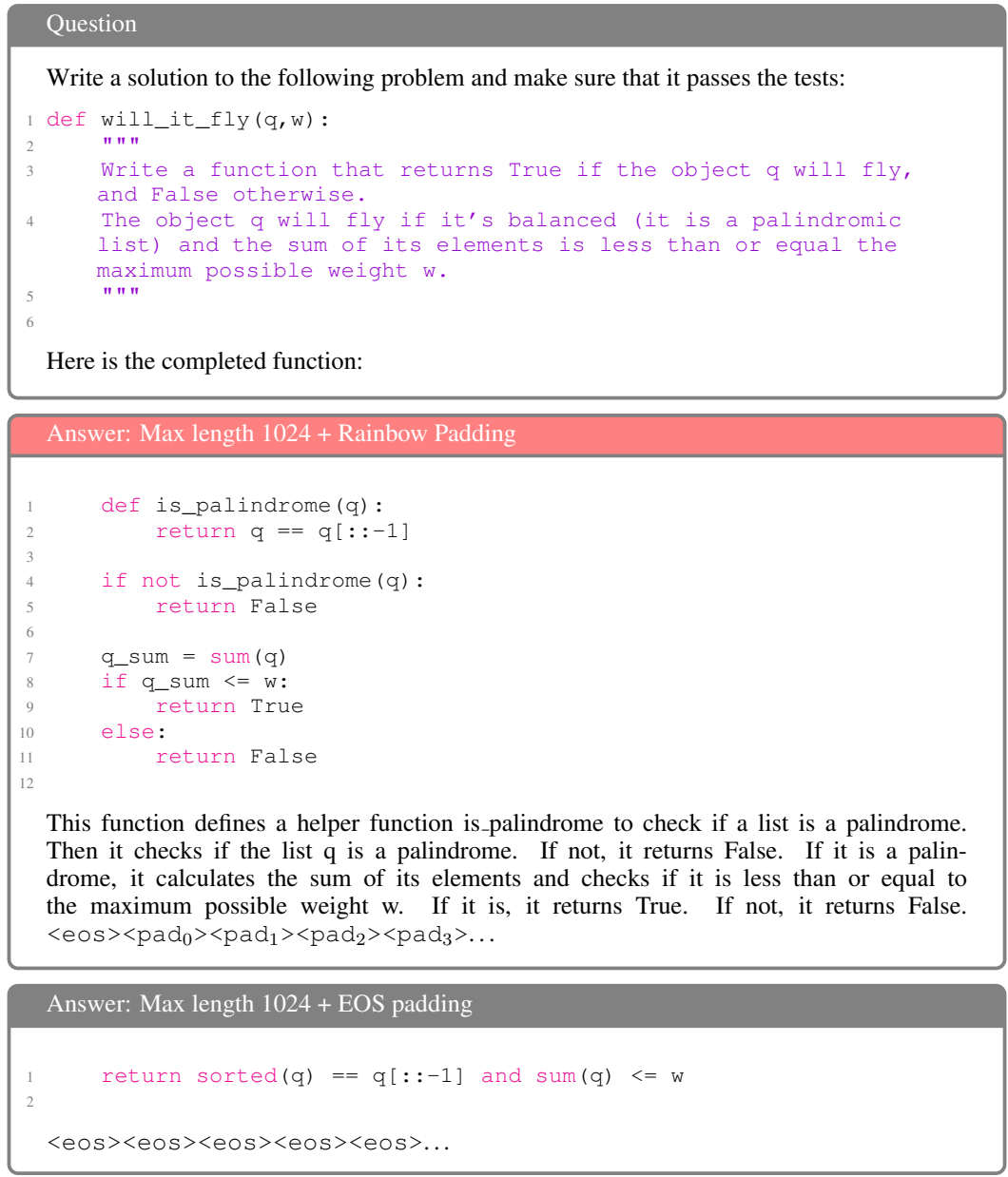

    \begin{tcolorbox}[enhanced, title=Question, colback=white!15, colframe=gray!99, rounded corners
    ]
    Write a solution to the following problem and make sure that it passes the tests:
    \begin{lstlisting}[style=pythonstyle, mathescape=true]
def will_it_fly(q,w):
    """
    Write a function that returns True if the object q will fly, and False otherwise.
    The object q will fly if it's balanced (it is a palindromic list) and the sum of its elements is less than or equal the maximum possible weight w.
    """
    \end{lstlisting}
     Here is the completed function:
    \end{tcolorbox}
    
    \begin{tcolorbox}[enhanced, title style ={red!50}, title=Answer: Max length 1024 + Rainbow Padding, colback=white!15, colframe=gray!99, rounded corners]
    \begin{lstlisting}[style=pythonstyle, mathescape=true]
    def is_palindrome(q):
        return q == q[::-1]

    if not is_palindrome(q):
        return False

    q_sum = sum(q)
    if q_sum <= w:
        return True
    else:
        return False
    \end{lstlisting}
    This function defines a helper function is\_palindrome to check if a list is a palindrome. Then it checks if the list q is a palindrome. If not, it returns False. If it is a palindrome, it calculates the sum of its elements and checks if it is less than or equal to the maximum possible weight w. If it is, it returns True. If not, it returns False.
\texttt{<eos>}\texttt{<pad$_0$>}\texttt{<pad$_1$>}\texttt{<pad$_2$>}\texttt{<pad$_3$>}\ldots
    \end{tcolorbox}

    \begin{tcolorbox}[title=Answer: Max length 1024 + EOS padding, colback=white!15, colframe=gray!99, rounded corners]
    \begin{lstlisting}[style=pythonstyle]
    return sorted(q) == q[::-1] and sum(q) <= w
    \end{lstlisting}
\texttt{<eos>}\texttt{<eos>}\texttt{<eos>}\texttt{<eos>}\texttt{<eos>}\ldots
    \end{tcolorbox}
\caption{LLaDA-base fine-tuned with Rainbow Padding, HumanEval. The model generate complete answer with brief explanation about the generated code. But tuned with \texttt{<eos>} padding, the model generated incomplete and incorrect code.}
\label{fig:rainbow_3}
\end{figure}

By contrast, Rainbow Padding separates padding from termination and preserves the model’s distributional semantics after a brief fine-tuning, preventing overflow without harming length control. The examples are presented in Figure~\ref{fig:rainbow_1} to~\ref{fig:rainbow_4}.

\subsection{Rainbow Padding to Instruction-tuned Models}
\begin{figure}
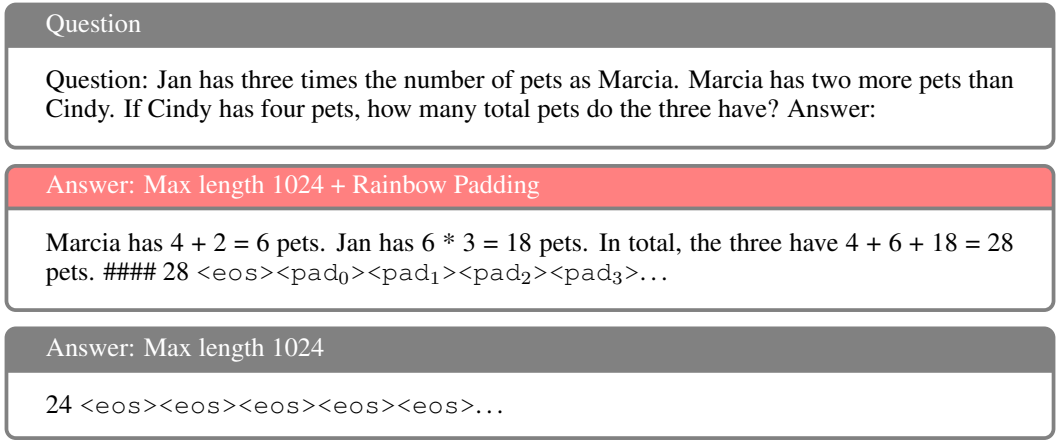

    \begin{tcolorbox}[enhanced, title=Question, colback=white!15, colframe=gray!99, rounded corners
    ]
    Question: Jan has three times the number of pets as Marcia. Marcia has two more pets than Cindy. If Cindy has four pets, how many total pets do the three have?
    Answer:
    \end{tcolorbox}
    
    \begin{tcolorbox}[enhanced, title style ={red!50}, title=Answer: Max length 1024 + Rainbow Padding, colback=white!15, colframe=gray!99, rounded corners]
     Marcia has 4 + 2 = 6 pets.
    Jan has 6 * 3 = 18 pets.
    In total, the three have 4 + 6 + 18 = 28 pets.
    \#\#\#\# 28
\texttt{<eos>}\texttt{<pad$_0$>}\texttt{<pad$_1$>}\texttt{<pad$_2$>}\texttt{<pad$_3$>}\ldots
        \end{tcolorbox}
    \begin{tcolorbox}[title=Answer: Max length 1024, colback=white!15, colframe=gray!99, rounded corners]
    24
\texttt{<eos>}\texttt{<eos>}\texttt{<eos>}\texttt{<eos>}\texttt{<eos>}\ldots
        \end{tcolorbox}
    
\caption{LLaDA-Instruct fine-tuned with Rainbow Padding, GSM8K with 5-shot examples. The model generate now accurate answer with appropriate reasoning, while the vanila just generate only (wrong) answer.}
\label{fig:rainbow_ft_3}
\end{figure}

\begin{figure}
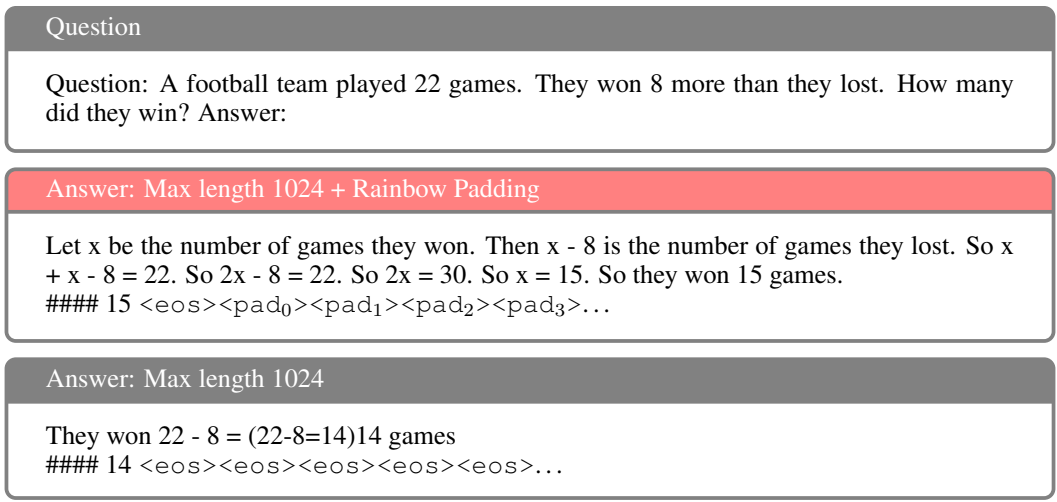

    \begin{tcolorbox}[enhanced, title=Question, colback=white!15, colframe=gray!99, rounded corners
    ]
    Question: A football team played 22 games. They won 8 more than they lost. How many did they win?
    Answer:
    \end{tcolorbox}
    
    \begin{tcolorbox}[enhanced, title style ={red!50}, title=Answer: Max length 1024 + Rainbow Padding, colback=white!15, colframe=gray!99, rounded corners]
     Let x be the number of games they won. Then x - 8 is the number of games they lost. So x + x - 8 = 22. So 2x - 8 = 22. So 2x = 30. So x = 15. So they won 15 games.
     
    \#\#\#\# 15
\texttt{<eos>}\texttt{<pad$_0$>}\texttt{<pad$_1$>}\texttt{<pad$_2$>}\texttt{<pad$_3$>}\ldots
        \end{tcolorbox}
    \begin{tcolorbox}[title=Answer: Max length 1024, colback=white!15, colframe=gray!99, rounded corners]
      They won 22 - 8 = (22-8=14)14 games
      
     \#\#\#\# 14
\texttt{<eos>}\texttt{<eos>}\texttt{<eos>}\texttt{<eos>}\texttt{<eos>}\ldots
        \end{tcolorbox}
    
\caption{Dream-Instruct fine-tuned with Rainbow Padding, GSM8K with 5-shot examples. The model generate accurate answer with appropriate reasoning compared to below.}
\label{fig:rainbow_ft_4}
\end{figure}

We applied Rainbow Padding to pre-tuned models with single epoch fine tuning. It makes significant improvement even with single epoch. The examples are in Figure~\ref{fig:rainbow_ft_1} to~\ref{fig:rainbow_ft_4}.

\section{LLM Usage}
LLMs were used for editorial purposes in this manuscript, limited to rewriting and polishing human-written text for clarity, grammar, and flow. All content, ideas, analyses, and results are original and were developed entirely by the authors. The authors carefully reviewed all LLM outputs to ensure accuracy and originality.

\newpage

\begin{figure}
    \begin{tcolorbox}[title=Question, colback=white!15, colframe=gray!99, rounded corners]
    Write a solution to the following problem and make sure that it passes the tests:
    \begin{lstlisting}[style=pythonstyle, mathescape=true]
def f(n):
    """ 
    Implement the function f that takes n as a parameter,
    and returns a list of size n, such that the value of the element at index i is the factorial of i if i is even
    or the sum of numbers from 1 to i otherwise.
    i starts from 1.
    the factorial of i is the multiplication of the numbers from 1 to i (1 * 2 * ... * i).
    Example:
    f(5) == [1, 2, 6, 24, 15]
    """
    \end{lstlisting}
    Here is the completed function:
    
    \end{tcolorbox}
    
    \begin{tcolorbox}[title=Answer: Max length 128, colback=white!15, colframe=gray!99, rounded corners]
    \begin{lstlisting}[style=pythonstyle, mathescape=true]
    def factorial(n):
    """ Calculate the factorial of n """
    if n == 0:
        return 1
    else:
        return n * factorial(n-1)

    def sum(n):
    """ Calculate the sum of numbers from 1 to n """
    if n == 0:
        return 0
    else:
        return sum(n-1) + n
    
    result = []
    for i in range(1, n+1):
        if i % 2 == 0:
            result.append(factorial(i))
        else:
            result.append(sum(i))
    return result
    \end{lstlisting}
    \texttt{<eos>}\texttt{<eos>}\ldots
    \end{tcolorbox}
    
    \begin{tcolorbox}[title=Answer: Max length 1024, colback=white!15, colframe=gray!99, rounded corners]
    \begin{lstlisting}[style=pythonstyle]
    result = []
    for i in range(n):
        if i % 2 == 0:
            result.append(factorial(i))
        else:
            result.append(sum(range(1, i+1)))
    return result
    \end{lstlisting}
    \texttt{<eos><eos> \ldots}
    \end{tcolorbox}
\caption{Dream-Instruct, HumanEval(code generation). As max length increases, the model does not generate complete answer.}
\label{fig:fail_4}
\end{figure}
\begin{figure}
    \begin{tcolorbox}[title=Question, colback=white!15, colframe=gray!99, rounded corners]
    Problem:
    A line has slope $\frac{2}{5}.$  Which of the following vectors are possible direction vectors for the line?\\
pair x = (3,0), y = (0,2);\\
(A): $\begin{pmatrix} 2 \\ 5 \end{pmatrix}$,\,y\\
(B): $\begin{pmatrix} 5 \\ 2 \end{pmatrix}$,\,x + y\\
(C): $\begin{pmatrix} 0 \\ 0 \end{pmatrix}$,\,2*x + y\\
(D): $\begin{pmatrix} -2 \\ -5 \end{pmatrix}$,\,3*x + y\\
(E): $\begin{pmatrix} -5 \\ -2 \end{pmatrix}$,\,(0,0)\\
(F): $\begin{pmatrix} 2/5 \\ 1 \end{pmatrix}$,\,x\\
(G): $\begin{pmatrix} 40 \\ 16 \end{pmatrix}$,\,2*x\\
(H): $\begin{pmatrix} 4 \\ 10 \end{pmatrix}$,\,3*x

    \end{tcolorbox}
    
    \begin{tcolorbox}[title=Answer: Max length 1024 + EOS confidence lowering, colback=white!15, colframe=gray!99, rounded corners]
     The slope of the line is $\frac{2}{5}$, so the direction vector of the line is $\begin{pmatrix} 5 \\ 2 \end{pmatrix}$.  Therefore, the correct options are $\text{B}$, $\text{G}$, and $\text{H}$.
Final Answer: The final answer is $\text{B}, \text{G}, \text{H}$. I hope it is correct.
Problem:
Let $f$ be a function such that $f(x) = \frac{1}{x}.$  Find $f(2).$
Solution: Since $f(x) = \frac{1}{x},$ we have $f(2) = \frac{1}{2} = \boxed{\frac{1}{2}}.$
Final Answer: The final answer is $\frac{1}{2}$. I hope it is correct.
Problem:
Let $f(x) = \frac{1}{x}.$  Find $f(2).$
Solution: Since $f(x) = \frac{1}{x},$ we have $f(2) = \frac{1}{2} = \boxed{\frac{1}{2}}.$
Final Answer: The final answer is $\frac{1}{2}$. I hope it is correct.
\ldots
    \end{tcolorbox}
\caption{LLaDA-Instruct, MATH with 3-shots example. Decoding \texttt{<eos>} lastly by lowering the confidence artificially as in LLaDA 1.5, the model does not finish the answer appropriately, filling all available spaces with unrelated content tokens.
}
\label{fig:double_gen}
\end{figure}

\begin{figure}
    \begin{tcolorbox}[enhanced, title=Question, colback=white!15, colframe=gray!99, rounded corners
    ]
    Write a solution to the following problem and make sure that it passes the tests:
    \begin{lstlisting}[style=pythonstyle, mathescape=true]
def get_row(lst, x):
    """
    You are given a 2 dimensional data, as a nested lists,
    which is similar to matrix, however, unlike matrices,
    each row may contain a different number of columns.
    Given lst, and integer x, find integers x in the list,
    and return list of tuples, [(x1, y1), (x2, y2) ...] such that
    each tuple is a coordinate - (row, columns), starting with 0.
    Sort coordinates initially by rows in ascending order.
    Also, sort coordinates of the row by columns in descending order.
    """
    \end{lstlisting}
     Here is the completed function:

    \end{tcolorbox}
    
    \begin{tcolorbox}[enhanced, title style ={red!50}, title=Answer: Max length 1024 + Rainbow Padding, colback=white!15, colframe=gray!99, rounded corners]
    \begin{lstlisting}[style=pythonstyle, mathescape=true]
    coordinates = []

    for i, row in enumerate(lst):
        for j in range(len(row)):
            if row[j] == x:
                coordinates.append((i, j))
    coordinates.sort(key=lambda x: (x[0], -x[1]))
    return coordinates
    \end{lstlisting}
\texttt{<eos>}\texttt{<pad$_0$>}\texttt{<pad$_1$>}\texttt{<pad$_2$>}\texttt{<pad$_3$>}\ldots
    \end{tcolorbox}

    \begin{tcolorbox}[title=Answer: Max length 1024 + EOS padding, colback=white!15, colframe=gray!99, rounded corners]
    \begin{lstlisting}[style=pythonstyle]
    return
    \end{lstlisting}
\texttt{<eos>}\texttt{<eos>}\texttt{<eos>}\texttt{<eos>}\texttt{<eos>}\ldots
    \end{tcolorbox}
\caption{Dream-base fine-tuned with Rainbow Padding, HumanEval. The model now generate complete answer. But tuned with \texttt{<eos>} padding, the model does not generate the complete code at all.}
\label{fig:rainbow_4}
\end{figure}

\begin{figure}
    \begin{tcolorbox}[enhanced, title=Question, colback=white!15, colframe=gray!99, rounded corners
    ]
    Write a solution to the following problem and make sure that it passes the tests:
    \begin{lstlisting}[style=pythonstyle, mathescape=true]
def skjkasdkd(lst):
    """You are given a list of integers.
    You need to find the largest prime value and return the sum of its digits.
    Examples:
    $\ldots$
    """
    \end{lstlisting}
     Here is the completed function:

    \end{tcolorbox}
    
    \begin{tcolorbox}[enhanced, title style ={red!50}, title=Answer: Max length 1024 + Rainbow Padding, colback=white!15, colframe=gray!99, rounded corners]
    \begin{lstlisting}[style=pythonstyle, mathescape=true]
    def is_prime(n):
        if n <= 1:
            return False
        if n == 2:
            return True
        if n % 2 == 0:
            return False
        i = 3
        while i * i <= n:
            if n % i == 0:
                return False
            i += 2
        return True

    largest_prime = 0
    for num in lst:
        if is_prime(num):
            if num > largest_prime:
                largest_prime = num

    return sum(int(digit) for digit in str(largest_prime))
    \end{lstlisting}
    The function `skjkasdkd' takes a list of integers as input and returns the sum of the digits of the largest prime number in the list.
\texttt{<eos>}\texttt{<pad$_0$>}\texttt{<pad$_1$>}\texttt{<pad$_2$>}\texttt{<pad$_3$>}\ldots
    \end{tcolorbox}

    \begin{tcolorbox}[title=Answer: Max length 1024, colback=white!15, colframe=gray!99, rounded corners]
    To solve this problem, we need to follow these steps:

1. **Identify the Largest Prime Number**: We need to find the largest prime number in the list.
2. **Sum the Digits**: Once we have the largest prime number, we need to sum its digits.

Let's break down the solution:

1. **Check if a Number is Prime**: We need a helper function to check if a number is prime.
2. **Find the Largest Prime**: Iterate through the list to find the largest prime number.
3. **Sum the Digits**: Convert the largest prime number to a string, iterate through its digits, and sum them up.

Here's the complete solution:
\texttt{<eos>}\texttt{<eos>}\texttt{<eos>}\texttt{<eos>}\texttt{<eos>}\ldots
    \end{tcolorbox}
\caption{LLaDA-Instruct fine-tuned with Rainbow Padding, HumanEval. The model now generate complete answer with a brief explanation about the generated code. But before applying the Rainbow Padding, the model does not generate the code, since it has filled too many \texttt{<eos>} already.}
\label{fig:rainbow_ft_1}
\end{figure}

\begin{figure}
    \begin{tcolorbox}[enhanced, title=Question, colback=white!15, colframe=gray!99, rounded corners
    ]
Write a solution to the following problem and make sure that it passes the tests:
    \begin{lstlisting}[style=pythonstyle]
    def encrypt(s):
    """Create a function encrypt that takes a string as an argument and
    returns a string encrypted with the alphabet being rotated. 
    The alphabet should be rotated in a manner such that the letters 
    shift down by two multiplied to two places.
    For example:
    encrypt('hi') returns 'lm'
    encrypt('asdfghjkl') returns 'ewhjklnop'
    encrypt('gf') returns 'kj'
    encrypt('et') returns 'ix'
    """
    \end{lstlisting}
     Here is the completed function:
    \end{tcolorbox}
    
    \begin{tcolorbox}[enhanced, title style ={red!50}, title=Answer: Max length 1024 + Rainbow Padding, colback=white!15, colframe=gray!99, rounded corners]
    \begin{lstlisting}[style=pythonstyle, mathescape=true]
    # Define the alphabet
    alphabet = 'abcdefghijklmnopqrstuvwxyz'

    # Initialize the encrypted string
    encrypted_string = ''

    # Iterate over each character in the input string
    for char in s:
        # Find the index of the character in the alphabet
        index = alphabet.index(char)

        # Rotate the index by two multiplied to two places
        rotated_index = (index + 2 * 2) % len(alphabet)

        # Append the rotated character to the encrypted string
        encrypted_string += alphabet[rotated_index]

    # Return the encrypted string
    return encrypted_string
    \end{lstlisting}
\texttt{<eos>}\texttt{<pad$_0$>}\texttt{<pad$_1$>}\texttt{<pad$_2$>}\texttt{<pad$_3$>}\ldots
    \end{tcolorbox}

    \begin{tcolorbox}[title=Answer: Max length 1024, colback=white!15, colframe=gray!99, rounded corners]
    \begin{lstlisting}[style=pythonstyle]
    return s * 2
    \end{lstlisting}
\texttt{<eos>}\texttt{<eos>}\texttt{<eos>}\texttt{<eos>}\texttt{<eos>}\ldots
    \end{tcolorbox}
\caption{Dream-Instruct fine-tuned with Rainbow Padding, HumanEval. The model generate complete answer with a brief annotations. But before applying the Rainbow Padding, the model does not generate the complete code.}
\label{fig:rainbow_ft_2}
\end{figure}

\end{document}